\documentclass[letterpaper, 12pt, journal, twoside]{IEEEtran}
\usepackage[top=57pt, left=48pt, right=60pt, bottom=54pt]{geometry}
\usepackage{algorithm}
\usepackage{algpseudocode}
\usepackage{amsmath, amssymb}
\usepackage{multirow}
\usepackage{cite}
\usepackage[pdftex]{graphicx}
\usepackage{url}
\usepackage{array}
\usepackage{booktabs}
\usepackage[table,xcdraw]{xcolor}
\usepackage{boldline}
\usepackage[utf8]{inputenc}
\usepackage[T1]{fontenc}
\usepackage[hidelinks]{hyperref}
\usepackage[caption=false,font=footnotesize]{subfig}
\usepackage{orcidlink}
\usepackage{tabularx}
\usepackage{ragged2e}
\usepackage{float} 
\usepackage{ragged2e}
\usepackage{tcolorbox}
\tcbuselibrary{skins,breakable}
\DeclareMathOperator*{\argmin}{arg\,min}

\newcommand{\FullTitle}{FISC: A Fluid-Inspired Framework for Decentralized and Scalable Swarm Control} 

\newcommand{\AuthorA}{Mohini Priya Kolluri}
\newcommand{\AuthorB}{Ammar Waheed} 
\newcommand{\AuthorC}{Zohaib Hasnain}

\newcommand{\AffilOne}{J. Mike Walker ’66 Department of Mechanical Engineering, Texas A\&M University, \\College Station, TX 77801, USA}
\newcommand{\AffilTwo}{Equal Contribution}
\newcommand{\AffilThree}{Corresponding Author (zhasnain@tamu.edu)}

\newcommand{\ORCIDA}{0009-0006-3829-6005}
\newcommand{\ORCIDB}{0000-0001-5042-6593}
\newcommand{\ORCIDC}{0000-0002-8358-451X}

\newcommand{\TFKeywords}{Swarm robotics; decentralized control; multi-agent systems; fluid-inspired primitives; computational fluid dynamics; collective motion.}

\newcommand{\TFRightBox}{%
\begin{tcolorbox}[
  enhanced,
  boxrule=0.6pt,
  colback=white,
  colframe=black!35,
  left=6pt,right=6pt,top=6pt,bottom=6pt,
  arc=0pt,
  width=\linewidth
]
{\bfseries Keywords}\par
\raggedright
\small \TFKeywords
\end{tcolorbox}%
}

\newcommand{\TFFirstPageStart}{%

{\LARGE \centering \bfseries \FullTitle\par}\vspace{8pt}

{\large \centering
\AuthorA$^{a,b}$,\ \AuthorB$^{a}$,\ \AuthorC$^{a,b,c}$\par
}\vspace{6pt}

{\normalsize \centering
$^{a}$\AffilOne
\vspace{6pt}\\
$^{b}$\AffilTwo  \quad \quad $^{c}$\AffilThree\par
}
\vspace{10pt}
\noindent
\begin{minipage}[t]{0.68\textwidth}
{\bfseries Abstract}\par\vspace{4pt} 
Achieving scalable coordination in large robotic swarms is often constrained by reliance on inter-agent communication, which introduces latency, bandwidth limitations, and vulnerability to failure. To address this gap, a decentralized approach for outer-loop control of large multi-agent systems based on the paradigm of how a fluid moves through a volume is proposed and evaluated. A relationship between fundamental fluidic element properties and individual robotic agent states is developed such that the corresponding swarm "flows" through a space, akin to a fluid when forced via a pressure boundary condition. By ascribing fluid-like properties to subsets of agents, the swarm evolves collectively while maintaining desirable structure and coherence without explicit communication of agent states within or outside of the swarm. The approach is evaluated using simulations involving $O(10^3)$ quadcopter agents and compared against Computational Fluid Dynamics (CFD) solutions for a converging-diverging domain. Quantitative agreement between swarm-derived and CFD fields is assessed using Root-Mean-Square Error (RMSE), yielding normalized errors of 0.15-0.9 for velocity, 0.61-0.98 for density, 0-0.937 for pressure. These results demonstrate the feasibility of treating large robotic swarms as continuum systems that retain the macroscopic structure derived from first principles, providing a basis for scalable and decentralized control.
\vspace{6pt}
\end{minipage}\hfill
\begin{minipage}[t]{0.3\textwidth}
\vspace{0pt}
\TFRightBox
\end{minipage}
\vspace{10pt}
}

\begin{document}
\twocolumn[
\TFFirstPageStart
]
\section{Introduction}
The study of swarm robotics has often been inspired by analogies to fluid systems \cite{oh2015survey}, yet these analogies remain largely metaphorical. Research in swarm robotics frequently borrows terms such as crowding pressure, alignment temperature, swarm density, and viscosity to describe emergent phenomena ranging from schooling fish to formations of aerial drones \cite{couzin2002collective, cucker2007emergent, sumpter2010collective, cavagna2010scale}. Despite their intuitive appeal, these descriptors are rarely grounded in a rigorous mathematical framework. The absence of well-defined swarm-level variables thus prevents the direct use of methods that are fundamental to fluid mechanics. 
\par
In most instances of classical fluid dynamics, four macroscopic variables (velocity $\mathbf{u}$, density $\rho$, pressure $P$, and temperature $T$) are sufficient to characterize many types of flow. These variables satisfy exact conservation laws, enable characteristic analysis of information transport, and admit constitutive relations consistent with thermodynamic principles \cite{anderson2003modern}. The equations of motion reduce to a closed set of balance laws that enable the prediction of flow evolution from specified initial and boundary conditions. In contrast, engineered robotic swarms lack an equivalent set of state variables. As a result, the formulations of fluid mechanics cannot be systematically applied to robotic swarms. Addressing this fundamental gap requires the identification of a minimal set of swarm observables that carry the same mathematical structure and exhibit physical relevance analogous to fluid primitive variables. 
\par
The work presented herein establishes a quartet \(Q:\{ \mathbf{u}_s, \rho_s, P_s, T_s\}\) of analogous primitive variables for swarm systems that maintain structure consistent with a Computational Fluid Dynamics (CFD) solution while accommodating discrete agent populations. Swarm velocity  $\mathbf{u}_s$ is derived from the mass-weighted projection of individual agents in the mission-defined drift direction. Swarm density $\rho_s$ is derived from the local agent concentration. Swarm pressure $P_s$ is defined from the acceleration-based momentum flux reflected through local velocity variance that is required to maintain coherence. Swarm temperature $T_s$ is derived from internal kinetic energy spread and available control authority. Constitutive parameters, such as specific heats and speed of sound, provide a foundation that bridges these primitive variables to fluidic properties enabling fluid-like swarm coordination. These definitions are designed to preserve conservation properties and allow translation of continuum fluid properties into distributed swarm behaviors. 
\par
By identifying and formalizing a minimal set of swarm-level primitive variables $Q$, this work establishes a principled foundation for treating large robotic swarms as continuum systems. The proposed quartet of variables is defined directly from agent-level states while preserving the structural role they play in classical fluid dynamics. This enables the direct transfer of conservation laws, constitutive relations, and characteristic flow behavior from compressible fluid theory into the swarm domain.
\par
The formulation presented herein advances fluid analogies in swarm robotics beyond descriptive metaphors by constructing variables that admit closure under balance laws at finite population sizes. A velocity-fitting procedure for a compressible flow field is introduced to transform these variables for decentralized control, embedding fluidic behavior implicitly into executable velocity commands without requiring inter-agent communication or physical boundaries. Together, these elements provide a scalable and analytically grounded framework for swarm coordination driven by continuum principles.

\section{Related Work}
Swarm motion (also referred to as collective motion in this study) research exhibits a clear methodological split along agent-based models and continuum descriptions. Agent-based traditions introduced by Reynolds's Boids model \cite{reynolds1987flocks, dehaan2009reynolds}, Vicsek's models \cite{vicsek1995novel, vicsek2012collective}, and Couzin's zone interaction rules \cite{couzin2002collective} capture individual stochasticity and anisotropic sensing but sacrifice the analytical tractability needed for formal guarantees. Hydrodynamic theories pioneered by Toner and Tu \cite{toner1998flocks} and extended by the active matter community enabled energy budgets, spectral cascades, and linear stability analysis, but they rely on ensemble averaging that removes agent addressability \cite{marchetti2013hydrodynamics}.
\par
More recent work has moved towards a thermodynamic description of active collectives. Haeri et al. \cite{haeri2021thermodynamics} derived swarm-level macroscopic states for attractive and repulsive particles and fit them to virial-style equations of state. The study confirmed that classical thermodynamic closures can persist in non-equilibrium swarms. In contrast to the presented work, these macroscopic states are not endowed with direct physical meaning.  
Viscido et al. \cite{viscido2024inferring} used high-speed video of fish schools to learn continuum balance laws directly from trajectories and validated that macroscopic conservation principles can be inferred from real biological collectives without explicit agent rules. Identifying canonical state variables that bridge microscopic dynamics and macroscopic conservation laws has been recently recognized as a grand challenge \cite{activematter2025roadmap}.
\par
Foundational work in distributed control and estimation formalized consensus and motion coordination over networked agents \cite{bullo2009distributed, katz2011distributed}, with extensions to UAV formations under switching topologies \cite{Zhou2022distributed} and decentralized surveillance with fixed-wing drones \cite{beard2006decentralized}. Conventional swarm robotics in centralized systems requires full state information of all swarm members to be available to all other members to maintain the desired swarm coherence/structure. While effective, these methods typically assume fixed interaction graphs or rely on structured communication, limiting scalability and flexibility in deformable or spatially adaptive formations. 
\par
Several frameworks based on Smoothed Particle Hydrodynamics (SPH) have modeled robot swarms as discretized fluid systems \cite{Mitikiri2021}. Pac et al. \cite{pac2007control} and Pimenta et al. \cite{pimenta2013sphswarm} modeled robot collectives as discretized fluids, producing emergent behaviors via local interactions. Recent systems such as FluidicSwarm \cite{eguchi2024fluidicswarm} and its obstacle-unaware variant \cite{eguchi2025fluidicnav} use SPH to shape formations or detect obstacles indirectly from velocity data. While these systems are behaviorally rich, they lack closed-form conservation laws or a minimal variable set that remains meaningful at finite swarm sizes.
\par
Notwithstanding their effectiveness, SPH-based and other similar approaches largely function as particle-level simulators, with fluidic behavior emerging through inter-agent potentials \cite{gazi2005stability}. Control is typically achieved through heuristic parameter tuning rather than principled mappings from continuum models. A recent survey \cite{yu2023overview} identifies key themes in swarm control such as formation, obstacle navigation, topology adaptation while emphasizing the lack of unifying, scalable, and physically grounded frameworks. Existing SPH and distributed methods either lack macroscopic structure guarantees or degrade in performance at finite agent populations \cite{romanczuk2012active, leonard2007collective, jadbabaie2003coordination}.  
\par
The literature highlighted above presents a gap between fluid-inspired descriptions of collective motion and swarm control frameworks that admit well-defined state variables, conservation structure, and direct use for decentralized control. Existing approaches either rely on agent-level interaction rules that lack macroscopic closure, derive continuum quantities only as descriptive diagnostics, or invoke mean-field limits that deteriorate at realistic swarm sizes. No existing paradigm provides a reciprocal mapping between agent variables and continuum fields. No existing work ascribes complete microscopic properties to macroscopic agents while retaining their macroscopic physical properties. Existing studies either insert fluid variables as post hoc diagnostics or derive them from mean field limits that degrade at realistic swarm sizes. As a result, fluid analogies in swarm robotics remain largely heuristic and do not provide a systematic basis for reconstructing macroscopic flow behavior or preserving volumetric structure in space. No existing framework has identified a minimal set of swarm observables that closes under conservation at finite population sizes, and no study has demonstrated real-time estimation and control on resource-limited robotic platforms.
\par
This work closes the gap by identifying a minimal quartet of swarm-level primitive variables that mirror the roles of classical fluid variables. These variables are defined from agent-level motion in a manner that supports conservation-consistent macroscopic behavior while remaining applicable to discrete, finite populations. The framework enables evaluation to address two questions: (i) Are the proposed variables sufficient to reconstruct any coherent macroscopic fluid-like behaviors? (ii) Does this formulation maintain volume structure similar to that of fluid flowing through the volume? 
\par
By introducing a velocity-fitting procedure derived from compressible flow fields, the formulation provides a concrete mechanism for translating continuum behavior into decentralized swarm motion, thereby extending fluid-inspired swarm models from descriptive analogies to principled, testable control constructs applicable across a wide range of robotic platforms. Furthermore, this work validates the theory through representative simulations involving a collection of agents representative of state-of-the-art low-cost robots. 
\par 
\textbf{The novelty of the approach lies in its ability to generate a trajectory solution for arbitrarily large collections of robots. This enables cohesive swarm motion through space without requiring any form of information sharing between agents or with external entities. Swarm features such as agent spacing, collective shape, and motion are governed by adjusting the swarm-equivalent fluid parameters and prescribed initial and boundary conditions. The framework retains agent-level granularity essential for feedback, sensing, and safety, while preserving the coordination structure that enables rigorous analysis of emergent order. The developed primitive variable definitions are implemented for a distributed swarm of aerial robots but are directly applicable to any collection of homogeneous or heterogeneous robotic agents.}

\section{Theoretical Framework}
A comprehensive nomenclature detailing all variables and symbols used in this work is provided in Section \ref{sec:nomenclature} for ease of reference. Throughout this text, the term \textit{agent} refers to an entity within the decentralized swarm framework, whereas \textit{robot} specifically represents a physical robotic platform (e.g., a quadcopter).

\subsection{Isentropic Flow Characteristics}
Isentropic flow is chosen for this analysis as it is well-understood and has closed-form analytical solutions for special cases. Here, it can provide a direct bridge between classical compressible fluid theory and collective swarm behavior. In thermodynamic gases, the isentropic assumption that combines adiabatic and reversible processes reduces the energy equation to a barotropic relation $P = k \rho^\gamma$, where $\gamma$ is the specific heat ratio and $k$ is the barotropic constant. The variables $P$, $\rho$, and $v$ are sufficient to close the governing \cite{anderson2003modern}. In swarm dynamics, an isentropic baseline assumes negligible communication loss, sensing error, and actuation dissipation. This regime isolates coordinative mechanics before realistic inefficiencies are introduced.
\par
The resulting equations admit a speed of sound analogue $c^2 = \partial P_s/\partial \rho_s$ that quantifies disturbance propagation through the swarm volume. Establishing this frictionless limit first yields scaling laws that will serve as reference solutions for future work when non-isentropic effects are included. Representative effects include communication delay, battery drain, and thermal noise. This separation clarifies which behaviors arise from fundamental coupling mechanisms versus dissipative processes.

\subsection{Flow Domain and Control Volume Formulation}
Fluid flow in a converging-diverging tunnel with a circular cross-section as shown in Fig. \ref{fig:domain} is considered as the reference solution for field-level swarm coordination. The domain spans $x\in[0,15]$ m with  inlet and outlet diameters of $3~m$ and the smallest diameter of $1.5~m$ is located at the axial section $x=6~m$ from the inlet. This domain is represented by a volume $V$ that is partitioned into several cubic sub-volumes $\Delta V_j$ with characteristic dimensions $\ell_{j}$, where $j$ is the index of the sub-volumes. A desired bulk drift velocity $\mathbf{u}_d(\mathbf{x},t)$ is provided as an external input and governs swarm motion. Movement control is realized by activating and deactivating sub-volumes to produce the desired propagation direction. Agent-level properties are computed through equivalence to continuum fluid flow under the given physical boundaries. Velocities are prescribed so that the desired drift velocity is set at the global level, while individual agent velocities inside each sub-volume follow from the fluid domain solution. Together with the state-variable mapping introduced later, this formulation enables fluid-like collective motion without requiring separate trajectory solutions for each agent. 

\section{Swarm-Domain Primitive Variables}
\subsection{Swarm Velocity Field: \texorpdfstring{$\mathbf{u}_s(\mathbf{x},t)$}{us(x,t)}}
Bulk motion of the swarm is modeled as one-dimensional along the desired drift velocity $\mathbf{u}_d(\mathbf{x},t)$, at position $x$ and time $t$. Let $\hat{u}_d$ denote the associated unit vector in the desired drift direction. Within a sub-volume $\Delta V_j$, consider $N$ agents with masses $m_i$ being commanded by velocity $\mathbf{v}_i$ where $i$ is the agent index. For purposes of this definition, it is assumed that the controller has perfect velocity tracking. The continuum velocity of a swarm within a $\Delta V_j$ is defined as the mass-weighted projection of agent velocity onto the drift direction;
\begin{equation}
\mathbf{u}_s(\mathbf{x},t) = \frac{1}{M(\mathbf{x},t)} \sum_{i \in \Delta V} m_i \, (\mathbf{v}_i \cdot \hat{u}_d)\,\hat{u}_d,
\label{eq:swarm_velocity}
\end{equation}
where the total mass $M(\mathbf{x},t)=\sum_{i \in \Delta V} m_i$ attempts to enforce momentum conservation. The dot product extracts the component of each agent motion that is parallel to the prescribed trajectory. Mass weighting ensures correct momentum aggregation in heterogeneous fleets. The resulting field is collinear with $\hat{u}_d$ and its magnitude measures the average advance speed along the intended path. Components of agent motion that are perpendicular to $\hat{u}_d$ are excluded from $\mathbf{u}_s$. These components are treated as fluctuations that contribute to temperature and pressure like variables introduced later, in the same way that transverse thermal agitation appears in thermodynamic properties rather than in the mean flow of a classical gas.
\par
This definition helps achieve two desired design features: \begin{enumerate} \item Momentum based construction ensures that the standard continuity statement holds exactly. The local time rate of change of mass density plus the divergence of mass flux is zero. This gives an exact coarse grained accounting of spatial agent distribution. \item The bulk velocity is constrained to a single mission defined axis. This recovers the simplest parallel with one-dimensional isentropic gas flow and allows direct use of barotropic pressure density linkage and acoustic speed relations. Motion that is perpendicular to the axis is treated as a fluctuation field. It contributes to the temperature and pressure variables in a manner similar to how random molecular agitation contributes to thermodynamic quantities in a classical gas.
\end{enumerate}

\subsection{Swarm Density: \texorpdfstring{$\rho_s(\mathbf{x},t)$}{rho\_s(x,t)}}
Within the swarm framework, density represents local concentration of coordinated mass rather than mere spatial occupancy. Swarm density $\rho_s(\mathbf{x},t)$ is defined as the drone concentration within a control volume $\Delta V_j = A_j\,\Delta x_j$, where $A$ is the effective cross-sectional area of a $\Delta V_j$ orthogonal to the drift direction and $\Delta x_j$ is a small axial slice along $\hat{u}_d$. For agents of identical mass $m_i$, define
\begin{equation}
\rho_s(\mathbf{x},t) = \frac{m_i N_j}{A_j \Delta x_j},
\label{eq:swarm_density}
\end{equation}
where $N_j$ is the number of agents that actively participate in coordinated motion inside $\Delta V_j$. This construction measures functional density that contributes to collective behavior. The variable area $A_j$ enables modeling through converging passages, diverging regions, and obstacles. The formulation remains tractable by treating flow as one-dimensional along the primary drift direction.
\par
The swarm density is the foundation for compressibility modeling in swarm dynamics. Under isentropic conditions, density variations follow a barotropic relation $P_s = k_s \rho_s^{\gamma_s}$. The parameter $\gamma_s$ is a swarm specific heat ratio that encodes compressibility and $k_s$ is the swarm-equivalent barotropic constant. Larger values indicate stronger resistance to density change. The relation permits compression and rarefaction behavior that is analogous to acoustic waves in gases. Regions with high agent concentration act as compression zones. Sparse regions act as rarefaction zones. A characteristic inter agent spacing $L_c$ follows from density as $L_c \approx \rho_s^{-1/3}$. This distance provides a design scale for interaction rules. It sets the spatial resolution of the continuum approximation and the effective sensing range used to synthesize responses to obstacles and to inter-agent interactions while preserving swarm coherence.

\subsection{Swarm Pressure: \texorpdfstring{$P_s(\mathbf{x},t)$}{Ps(x,t)}}
Swarm pressure arises partly from inter-agent momentum exchange and mostly from acceleration resulting from changing velocity commands needed for fluid-like movement and structure in space. It does not arise from random molecular collisions as in a classical fluid. It quantifies the normal acceleration field that would be needed to completely reverse agent momentum at the boundary of a control volume. Therefore, it can be used to determine how the agents should move within a sub-volume such that it maintains shape.  
\par
Swarm pressure is defined through an acceleration-based momentum flux analysis extending classical kinetic theory to coordinated swarms. Consider a cubic sub-volume with edge length $\ell$, volume $\Delta V=\ell^3$, and face area $A=\ell^2$. For agents with masses $m_i$ and velocities $\mathbf{v}_i$ contained within the sub-volume, pressure contribution at each face emerges from acceleration required to reverse velocity components normal to that boundary.

\subsubsection{Normal Stresses and Isotropic Swarm Pressure}
For the positive $x$-face with outward unit normal $\hat{n}_x = (1,0,0)$, each agent contributes a velocity component $v_{x,i}=\mathbf{v}_i\cdot \hat{n}_x$. Momentum exchange required for velocity reversal generates impulse $\Delta P_i = 2\,m_i\,v_{x,i}$ over characteristic reversal time scale $t_{i,\text{rev}} = \Delta x/|v_{x,i}|$, representing uniform deceleration across half the cell dimension. This yields an average force density contribution on each face:
\begin{equation}
F_{x,i} = \frac{\Delta P_i}{t_{i,\text{rev}}} = \frac{2\,m_i\,v_{x,i}^2}{\Delta x}
\end{equation}
The normal stress component on the $x$-face becomes:
\begin{equation}
P_{xx} = \frac{1}{A}\sum_{i\in \Delta V} F_{x,i} = \frac{2}{\Delta V}\sum_{i\in \Delta V} m_i\,v_{x,i}^{2}
\end{equation}
Extending this analysis to all three coordinate directions yields stress tensor components:
\begin{align}
P_{xx}&=\frac{2}{\Delta V}\sum_{i\in \Delta V} m_i v_{x,i}^{2}, \nonumber\\
P_{yy}&=\frac{2}{\Delta V}\sum_{i\in \Delta V} m_i v_{y,i}^{2}, \\
P_{zz}&=\frac{2}{\Delta V}\sum_{i\in \Delta V} m_i v_{z,i}^{2}. \nonumber
\end{align}
The isotropic swarm pressure is the trace of this momentum exchange tensor:
\begin{align}
P_s &= \frac{2*(P_{xx}+P_{yy}+P_{zz})}{3} \nonumber \\
    &= \frac{2}{3\Delta V}\sum_{i\in \Delta V} m_i\!\left(v_{x,i}^{2}+v_{y,i}^{2}+v_{z,i}^{2}\right) \\
    &= \frac{2}{3\Delta V}\sum_{i\in \Delta V} m_i\,\|\mathbf{v}_i\|^{2} \nonumber
\label{eq:swarm_pressure}
\end{align}
Expressing this using swarm density $\rho_s=\frac{1}{\Delta V}\sum_{i\in \Delta V} m_i$ and mean-square velocity $\langle \|\mathbf{v}_i\|^{2}\rangle$:
\begin{equation}
P_s=\frac{2}{3}\,\rho_s\,\langle \|\mathbf{v}_i\|^{2}\rangle
\label{eq:pressure_density}
\end{equation}
To isolate internal swarm coordination pressure from bulk drift, the velocities relative to the local swarm velocity $\mathbf{u}_s$ are decomposed:
\begin{equation}
P_{s,\text{internal}}=\frac{2}{3}\,\rho_s\,\langle \|\mathbf{v}_i-\mathbf{u}_s\|^{2}\rangle
\label{eq:internal_pressure}
\end{equation}
This internal pressure captures velocity variance that drives the temperature variable in the swarm thermodynamic formulation and links kinetic disorder to mechanical stress.
\begin{figure*}
    \centering
    {\includegraphics[width=1\textwidth]{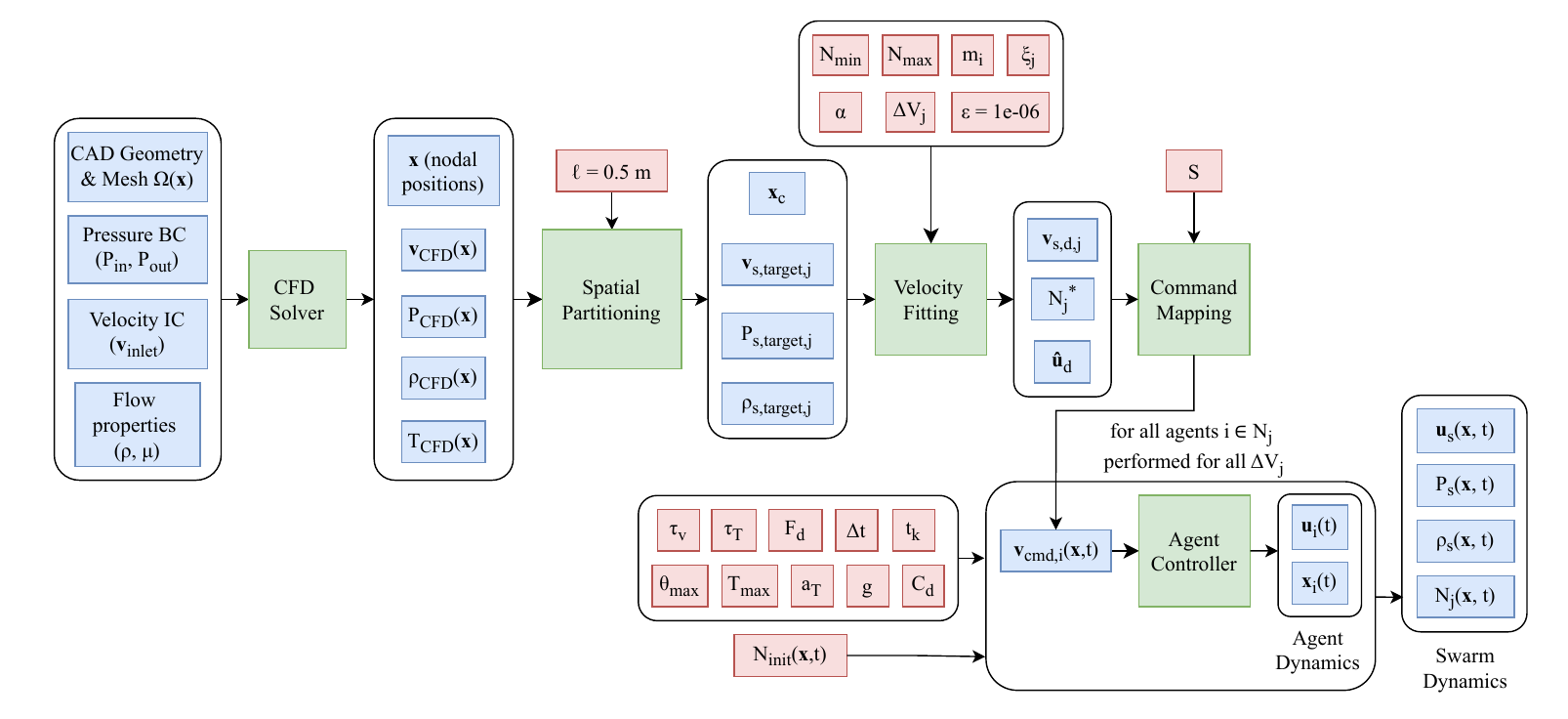}} \\
    \caption{Flow diagram showing evolution of field variables in the FISC framework.}
    \label{fig:Flowdiagram}
\end{figure*}

\subsection{Swarm Temperature Field: \texorpdfstring{$T_s(\mathbf{x},t)$}{Ts(x,t)}}
The proposed formulation of swarm temperature hypothesizes  that it measures internal kinetic disorder within a control volume, i.e., it quantifies how individual agent velocities deviate from the local mean motion. Unlike thermal temperature in gases that reflect random molecular agitation, swarm temperature captures variance in coordination. This definition is devised to (i) capture local random kinetic energy fluctuations and (ii) encode control authority for robust disturbance response. 
\begin{equation}
T_s = T_{s,\text{rand}} + T_{s,\text{ctrl}} = \frac{1}{c_{v,s}}\bigl(e_{\text{rand}} + \lambda\, e_{\text{ctrl}}\bigr)
\label{eq:swarm_temperature}
\end{equation}
The swarm-specific heat $c_{v,s}$ converts energy to temperature, analogous to specific heat at constant volume for fluids. Here, $e_{\text{rand}}=\tfrac{1}{2} \|\mathbf{v}_i-\mathbf{U}\|^{2}$ is random kinetic energy per unit mass. The random fluctuation component $T_{s,\text{rand}}$ in Eq.\eqref{eq:temperature_random} uses the velocity variance relative to the local mean.
\begin{equation}
T_{s,\text{rand}} = K_{b,s}\,\frac{\displaystyle \sum_{i\in \Delta V} m_i\,\|\mathbf{v}_i-\mathbf{U}\|^{2}}{2\,\displaystyle \sum_{i} m_i},
\label{eq:temperature_random}
\end{equation}
where $\mathbf{U}=\frac{\sum_{i\in \Delta V} m_i\,\mathbf{v}_i}{\sum_i m_i}$ is the local mass-weighted mean velocity. The constant $K_{b,s}$ is a scaling factor that is devised to be analogous to Boltzmann’s constant that maps velocity variance to swarm energy units.
\par
The $T_{s,\text{ctrl}}$ in Eq. \eqref{eq:swarm_temperature} explicitly incorporates control authority with sensing and actuation latency with deliberate behavioral diversity for disturbance response. The available control authority is associated with the swarm analogue of speed of sound $c$ which represents the propagation speed of coordinated disturbances of the robots. Here $e_{\text{ctrl}}=a_{\max}\,L_c$ is the available control energy, where $a_{\max}=F_{\max}/m$ is the maximum specific acceleration and $0<\lambda<1$ sets aggressiveness in the use of the drive budget. This formulation for temperature enables a relationship between temperature and disturbance propagation, which is similar to that within the body of a physical fluid, where $c=\sqrt{\gamma_s R_s T_s}$, implying higher temperatures corresponding to faster propagation. For the swarm domain, a larger value of $T_s$, with all other factors being identical, indicates the presence of more control authority and a faster rate of disturbance propagation.  

\subsection{Constitutive Parameters}
The following constitutive parameters are formulated to establish relationships between the primitive variables of swarms analogous to those in compressible fluids. While these parameters are not directly applicable to the current control volume assignment framework, their formulation clarifies the theoretical foundation and identifies future extensions. 
\paragraph{Specific heats $c_{p,s}$ and $c_{v,s}$}
The constant-volume specific heat $c_{v,s}$ maps swarm temperature to internal energy per unit mass as $e_s=c_{v,s} T_s$. The specific heat $c_{p,s}$ maps swarm temperature to its enthalpy as $h_s = e_s + P_s/\rho_s = c_{p,s}\,T_s$.
\paragraph{Specific heat ratio $\gamma_s$}
The ratio $\gamma_s=c_{p,s}/c_{v,s}$ quantifies swarm compressibility. Under isentropic conditions the closure is $P_s = C \rho_s^{\gamma_s}$. Larger $\gamma_s$ indicates a less compressible collective and a higher control effort to sustain formation during density change.
\paragraph{Gas constant $R_s$}
The swarm gas constant definition $R_s = c_{p,s}-c_{v,s}$ is established with state relation $P_s=\rho_s R_s T_s$. This parameter captures added disorder during compression at constant pressure versus constant entropy. It sets sensitivity to agitation and influences the static modulus. 
\paragraph{Speed of sound $c$}
The swarm speed of sound represents propagation velocity of coordinated disturbances through collective motion, analogous to acoustic wave speed in compressible media. It sets time scales for formation change, obstacle response, and mission updates.  An effective speed of sound formulation in Eq. \eqref{eq:sound_speed} includes reaction delay $\tau$  that arises from actuation latency, consensus filtering, or controller buffering and angular frequency of a disturbance $\omega$.
\begin{equation}
c^{2}=\frac{\gamma_s\,R_s\,T_s}{1+\omega^{2}\tau^{2}},
\label{eq:sound_speed}
\end{equation}
This formulation captures the trade-off between response speed and stability. High-frequency disturbances experience reduced propagation speed due to finite reaction time. When reaction delay $\tau$ does not scale proportionally with $T_s$, high frequency disturbance propagation may experience dispersion effects as agents expend energy on random motion rather than coordinated response. At low frequencies ($\omega \ll 1/\tau$), the speed simplifies to $c = \sqrt{\gamma_s R_s T_s}$. The temperature decomposition into random and control components directly affects sound speed through available control authority. 
\begin{figure}
\centering
\includegraphics[width=1\linewidth]{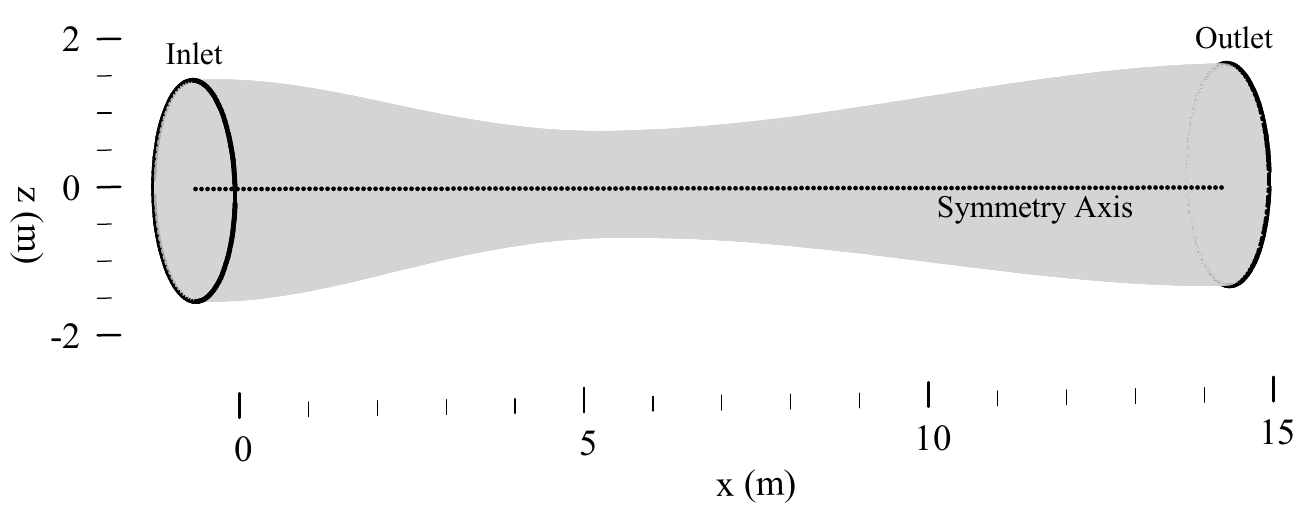}
\caption{Axisymmetric CD Nozzle in XZ plane}
\label{fig:domain}
\end{figure}

\section{Implementation of Theoretical Framework}
This section covers the implementation methodology, which includes spatial domain partitioning and velocity-fitting. A Computational Fluid Dynamics (CFD) simulation of airflow through a converging-diverging nozzle provides the reference solution for field-level swarm coordination. This implementation methodology facilitates swarm domain discretization and control command mapping that enables the desired fluid behavior to be mapped akin to the CFD fields.

\subsection{Implementation methodology background}
A velocity-only control architecture is assumed for the swarm, which is consistent with the actuation and sensing capabilities of low-cost aerial platforms envisioned for future experimental validation. Enforcing simultaneous pressure and velocity constraints would require control authority over higher-order temporal derivatives, which is generally infeasible for such systems.
\par
In the CFD solution, the pressure and velocity fields are inherently coupled through the governing equations of fluid motion. Accordingly, if a velocity field can be identified within the swarm domain that satisfies an equivalent coupling relationship, the emergent swarm dynamics are expected to approximate the behavior of the corresponding fluid. Two principal aspects of this analogy are: (1) the preservation of the shape of individual fluid elements, corresponding to structural cohesion within the swarm, and (2) the translational motion of these elements corresponding to the swarm’s collective kinematic response.
\par
The objective of this experimentation is to derive swarm-level fields from fluid solution fields for pressure and velocity primitive variables. The swarm-domain formulations for pressure and velocity incorporate parameters for mass and density that do not possess direct physical analogues in the fluid domain. Consequently, the total agent mass contained within a swarm sub-volume is expected to differ from the mass of the corresponding fluid element. Similarly, although a swarm density definition can be established for analytical purposes, it may not be strictly analogous to the density of a continuous fluid. Nevertheless, by careful initial selection of these parameters, it is possible to achieve aggregate swarm behavior that reproduces the desired velocity-pressure characteristics of the reference flow field. This correspondence is achieved through the application of an appropriate velocity-fitting procedure \cite{nocedal2006numerical}. 
\par
A smooth converging-diverging tunnel geometry was chosen to establish whether the agent motion mirrors fluid flow through such a structured domain. This tests the framework's ability to reproduce compressible effects in agent motion, such as acceleration, compression, and expansion without the explicit definition of physical boundaries. 

\subsection{Spatial Partitioning}
The converging-diverging tunnel domain is uniformly partitioned into cubic control sub-volumes with edge length of $\ell = 0.5$ m. The control volume averaging is carried out by utilizing the velocity $\mathbf{v}_{\text{CFD}}$ and pressure $P_{\text{CFD}}$ fields from the CFD flow solution to compute their averaged values at each of these control volumes. Input data consist of spatially varying CFD-derived velocity and pressure fields at all nodal positions $\{\mathbf{x}, \mathbf{v}_{\text{CFD}}, P_{\text{CFD}}\}$ obtained from steady-state Ansys Fluent simulations of a converging-diverging tunnel. For each control volume centered at $\mathbf{x}_c = (x_c, y_c, z_c)$, the target velocity $\mathbf{{v}_{\text{s,target,j}}}(\mathbf{x}_c)$ and pressure $P_{\text{s,target,j}}(\mathbf{x}_c)$ values are computed via spatial averaging of the respective CFD solution values $\mathbf{v}_{\text{CFD}}(\mathbf{x})$ and $P_{\text{CFD}}(\mathbf{x})$ at all spatial nodes $\mathbf{x} = (x,y,z)$ for all $\Delta V_j$. 

\subsection{Velocity-Fitting}
The CFD solution provides spatially varying target fields of velocity $\mathbf{v}_{\text{CFD}}$ and pressure $P_{\text{CFD}}$. These are averaged over each control volume $\Delta V_j$ to yield local targets of velocity $\mathbf{v}_{\text{s,target,j}}(\mathbf{x}_c)$ and pressure $P_{\text{s,target,j}}(\mathbf{x}_c)$. The computed target values are utilized to derive the swarm-level velocity field through velocity-fitting. Velocity-fitting derives swarm-feasible velocity commands $\textbf{v}_{s,d,j}$ while constraining agent densities (agent count $N_j$ per CV) from CFD target $\mathbf{{v}_{\text{s,target,j}}}$ and $P_{\text{s,target,j}}$ fields while preserving fluid physics. It addresses the transition from continuous fluid fields to discrete swarm systems by scaling velocities to match agent capabilities and constraining densities to a realistic number of agents per sub-volume. 
\par
Specifically, the velocity-fitting performs an optimization for each $\Delta V_j$ to determine the agent density $N_j^*$  (agent count per control sub-volume) and swarm velocities $\mathbf{v}_{s,d,j}$ within each control volume that reproduce the commanded velocities close to the target CFD velocities while respecting macroscopic feasibility. 
\par
Given the target velocity $\mathbf{v}_{\text{s,target,j}}(\mathbf{x_c})$ and pressure $P_{\text{s,target,j}}(\mathbf{x_c})$ from CFD, the following coupled system is enforced to obtain the desired swarm velocities $\mathbf{v}_{s,d,j}$ for each $\Delta V$ of index $j$:
\begin{align}
\mathbf{v}_{s,target,j} &= \frac{1}{N_j}\sum_{i=1}^{N_j} \mathbf{v}_{s,d,j}, \label{eq:velocity_constraint}\\
P_{\text{s,target,j}} &= \frac{2m_i}{3\Delta V_j}\sum_{i=1}^{N_j} \|\mathbf{v}_{s,d,j} - \mathbf{v}_{s,target,j}\|^2, \label{eq:pressure_constraint}
\end{align}
where $m_i$ is the individual-agent mass. Eq. \eqref{eq:velocity_constraint} constrains the agent velocities to average to the target velocity field. Eq. \eqref{eq:pressure_constraint} enforces that the kinetic energy associated with velocity fluctuations produces the desired pressure field. The agent count in a sub-volume $N_j$ influences both mean and variance terms, introducing a coupling between discrete (agent count) and continuous (velocity vectors) variables. Realistic density bounds are imposed by the inequality constraint $N_{\min} \leq N_j \leq N_{\max}$. Forcing $N_j$ between $N_{min} =2$ and $N_{max} = 10$ ensures that the agent distribution remains physically feasible, enabling pressure preservation under velocity changes. 
\par
The velocity-fitting is executed independently for each sub-volume. The solver computes an optimized $N_j^*$ and corresponding desired velocities $\mathbf{v}_{s,d,j}$ that satisfy the coupled constraints in Eqs. \eqref{eq:velocity_constraint} and \eqref{eq:pressure_constraint}. Convergence is assessed using a fixed tolerance $\epsilon = 10^{-6}$. Validation is performed using residual metrics. Velocity and pressure accuracies are quantified by their Root Mean Square Error ($\text{RMSE}$) values. These are computed over the set of valid control sub-volumes, denoted $N_j$. 
The derived swarm-level velocity commands through velocity-fitting are then utilized as control commands for the agents based on their spatial locations in the tunnel.
\par
The swarm density scaling problem is addressed by treating $N_j$ as a discrete decision variable and solving a continuous subproblem for each candidate value. For fixed $N_j \in [N_{\min}, N_{\max}]$, the objective function in Eq. \eqref{eq:iterative_objective} is minimized, where $\bar{\mathbf{v}}_j = \frac{1}{N_j} \sum_{i=1}^{N_j} \mathbf{v}_i$, and $\alpha$ is a scalar weighting parameter that balances fidelity to velocity and pressure targets. Each $N_j$ subproblem is solved using a Newton-Raphson iteration scheme with the update direction computed using the Jacobian.
\begin{equation}
\begin{split}
\min_{\{\mathbf{v}_{s,d,j}\}} 
&\left\|\mathbf{v}_{\text{s,target,j}} - \frac{1}{N_j} \sum_{i=1}^{N_j} \mathbf{v}_{s,d,j} \right\|^2 \\
&+ \alpha \left| P_{\text{s,target,j}} - 
\frac{2m_i}{3\Delta V_j} \sum_{i=1}^{N_j} 
\|\mathbf{v}_{s,d,j} - \bar{\mathbf{v}}_j \|^2 \right|,
\end{split}
\label{eq:iterative_objective}
\end{equation}
Initialization is performed using perturbed target velocities:
\begin{equation}
\mathbf{v}_j^{(0)} = \mathbf{v}_{\text{s,target,j}} + \sigma \boldsymbol{\xi}_j, \quad \boldsymbol{\xi}_j \sim \mathcal{N}(\mathbf{0}, \mathbf{I}),
\end{equation}
where $\sigma^2 \approx P_{\text{s,target}} (3\Delta V)/(2m_i N_j)$ is estimated from the pressure target. Iterations continue until the update norm satisfies $\|\Delta\mathbf{v}^{(k)}\| < \epsilon$. The optimal agent count is selected by minimizing the residual from the objective:
\begin{equation}
N_j^* = \argmin_{N_j \in [N_{\min}, N_{\max}]} \mathcal{L}(N_j),
\end{equation}
where $\mathcal{L}(N_j)$ denotes the final loss value for each candidate $N_j$. This enumeration guarantees the global minimum within the discrete range at a computational cost of $\mathcal{O}((N_{\max} - N_{\min}) \cdot K_{\text{iter}} \cdot N_j)$, where $K_{\text{iter}}$ is the average number of iterations per candidate $N_j$.

\section{Individual-Agent Velocity Plant}
The velocity controller plant \textit{VelocityPlant} maps commanded body velocities to realized body velocities for each quadrotor agent. It abstracts the influence of inner-loop attitude control and thrust modulation under aerodynamic loading. The resulting time series are subsequently aggregated into swarm-level fields. All quantities are defined in the body-fixed frame, with $x$ forward, $y$ right, and $z$ downward, following the North-East-Down (NED) convention. This plant model is adopted for large-scale simulations where repeatable tracking and moderate computational cost are required. The fidelity is sufficient for constructing swarm fields and for evaluating control laws at the collective scale.

\subsection{State Input, Output and Internal Structure}
The plant state comprises the body velocity $\mathbf{u}_i = [u\ v\ w]^\top$ and the realized thrust acceleration $\mathbf{a}_T$. The input is the commanded body velocity $\mathbf{v}_{\text{cmd,i}}$. The output is the realized velocity $\mathbf{v}$. The translational dynamics of the vehicle are given by:
\begin{equation}
\dot{\mathbf{v}} = \mathbf{a}_T + \begin{bmatrix}0 \\ 0 \\ g \end{bmatrix} + \frac{\mathbf{F}_d}{m_i},
\end{equation}
where $m$ is the vehicle mass, $g$ is gravitational acceleration, and $\mathbf{F}_d$ is the aerodynamic drag force computed per axis using a quadratic airspeed model. The thrust acceleration follows a first-order lag:
\begin{equation}
\dot{\mathbf{a}}_T = \frac{\mathbf{a}_{T,\text{d}} - \mathbf{a}_T}{\tau_T},
\end{equation}
with time constant $\tau_T = 0.08s$. The desired thrust $\mathbf{a}_{T,\text{d}}$ is formed by combining proportional shaping on velocity error, gravity compensation, and an optional feedforward term derived from estimated aerodynamic drag at the commanded velocity.

\subsection{Actuation Limits and Constraint Handling}
Thrust commands are constrained prior to actuation and the lateral motion is limited by a tilt cone defined by the maximum tilt angle $\theta_{\max}$. The total thrust magnitude is bounded by $T_{\max}$. These constraints ensure feasible trajectories and prevent violations of vehicle's dynamic envelope during aggressive inputs.

\subsection{Numerical Integration}
Time integration proceeds on a strictly increasing temporal grid. A zero-order hold is applied to $\mathbf{v}_{\text{cmd}}$ over each interval. The plant uses a semi-implicit Euler method for updates, with optional sub-stepping to maintain stability when $\tau_T$ is small or when input commands vary abruptly. The method remains stable across all operational conditions considered in swarm-scale studies.

\subsection{Nominal Parameters}
The nominal configuration of an agent is modeled as a small quadrotor with mass $m_i = 1$kg and a thrust-to-weight ratio $T_{\max}/(mg) = 2.2$. The air density is set to $\rho = 1.225$ kg\,m$^{-3}$. Reference areas are $\mathbf{S} = [0.02\ 0.02\ 0.03]^\top$ m$^2$, and the drag coefficients are $\mathbf{C}d = [1.0\ 1.0\ 1.2]^\top$. The maximum tilt angle is $\theta_{\max} = 55^\circ$. The shaping constant is $\tau_v = 0.5$ s, and the inner lag time is $\tau_T = 0.08$ s. 

\subsection{Performance Testing Prior to Swarm Use}
\begin{figure}
\centering
\includegraphics[width=1\linewidth]{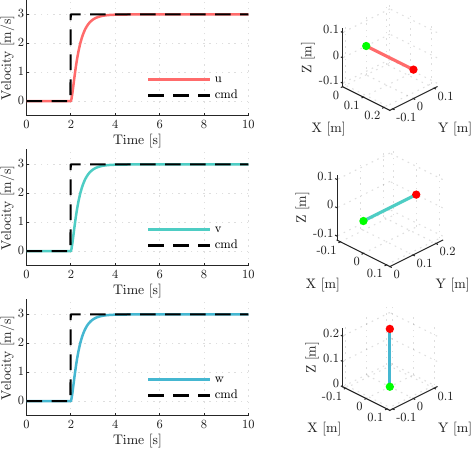}
\caption{Velocity tracking and trajectory plots for step responses of individual axes.}
\label{fig:single_axis}
\end{figure}

\begin{figure}
\centering
\includegraphics[width=1\linewidth]{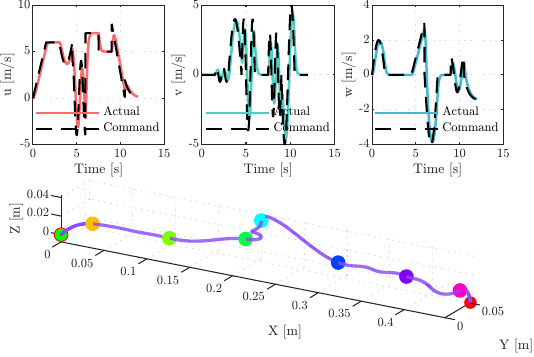}
\caption{Aggressive maneuvers simulation demonstrating effect of feedforward gain on velocity tracking performance on different racing maneuvers.}
\label{fig:race}
\end{figure}

\noindent Prior to integration with swarm simulations, the plant's performance was assessed using a structured test suite. Tests evaluated functional behavior, hover stability, single-axis and multi-axis step tracking, drag sensitivity, and the effectiveness of drag feedforward. Additional scenarios included parameter sweeps, thrust-to-weight variation, constraint enforcement, frequency response, cross-axis coupling, wind disturbance rejection, and robustness to sensor noise. Acceptance criteria were defined in terms of tracking accuracy, constraint satisfaction, and axis decoupling within the expected operating envelope.
\begin{itemize}
\item Single-axis step tests reached the commanded value with very minimal steady state error. Settling time was about $1.67$ s with no overshoot as shown in Fig. \ref{fig:single_axis}.
\item Thrust to weight sweeps from $1.5$ to $6.0$ showed a common maximum speed near $7.48$ m s$^{-1}$. Peak tilt approached $54.3^\circ$ and identified attitude as the active constraint.
\item Constant headwind from zero to $8$ m s$^{-1}$ with feedforward gain $0.8$ produced steady state errors between $0.01$ and $0.06$ m s$^{-1}$.
\item Monte Carlo noise tests with velocity noise from zero to $0.5$ m s$^{-1}$ RMS produced RMSE that rose from about $0.57$ to about $2.98$ m s$^{-1}$ on the lateral axes and from about $0.19$ to about $2.18$ m s$^{-1}$ on the vertical axis.
\item The aggressive simulation with high-speed maneuvers as shown in Fig. \ref{fig:race} shows the plant tracking capabilities in all three directions.
\end{itemize}

The plant exhibits behavior that is dynamically similar to that of a low-cost commercially available quadcopter. This was a key requirement prior to implementation in swarm configuration as it enables future hardware validation.
\begin{figure*}
\centering
{\includegraphics[width=0.49\textwidth]{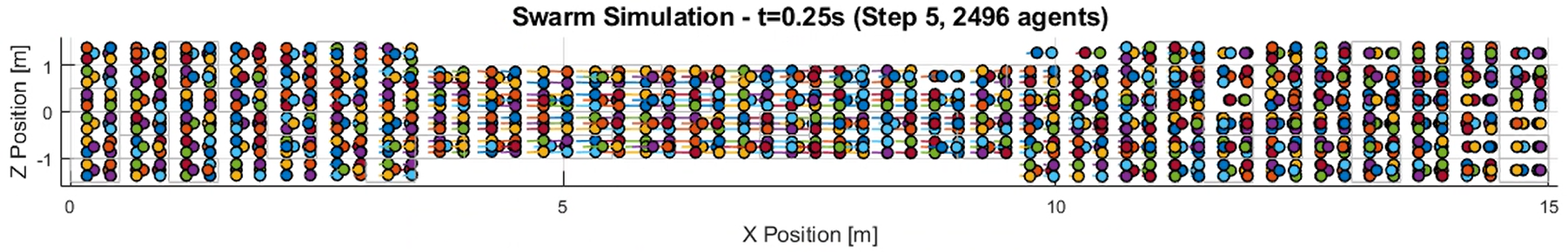}} \hfill
{\includegraphics[width=0.49\textwidth]{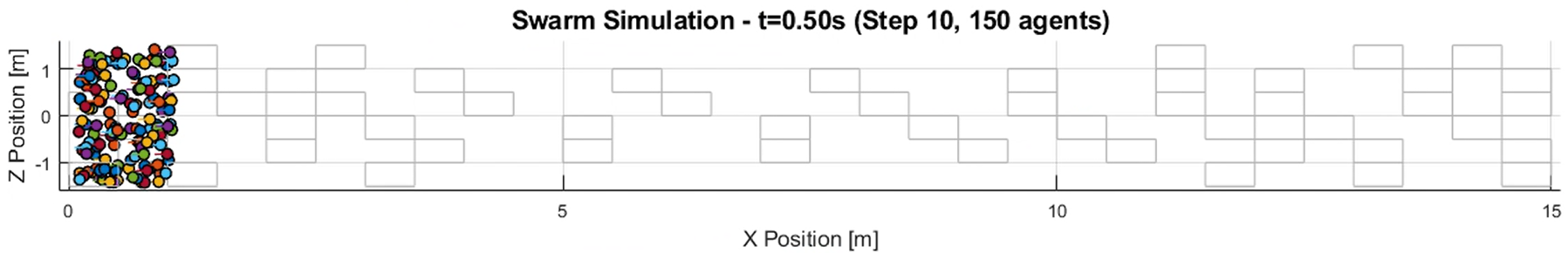}} \\
{\includegraphics[width=0.49\textwidth]{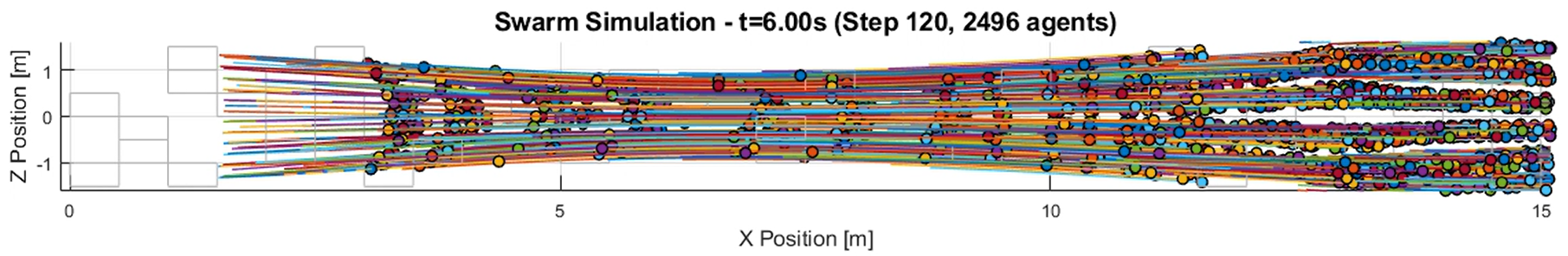}} \hfill
{\includegraphics[width=0.49\textwidth]{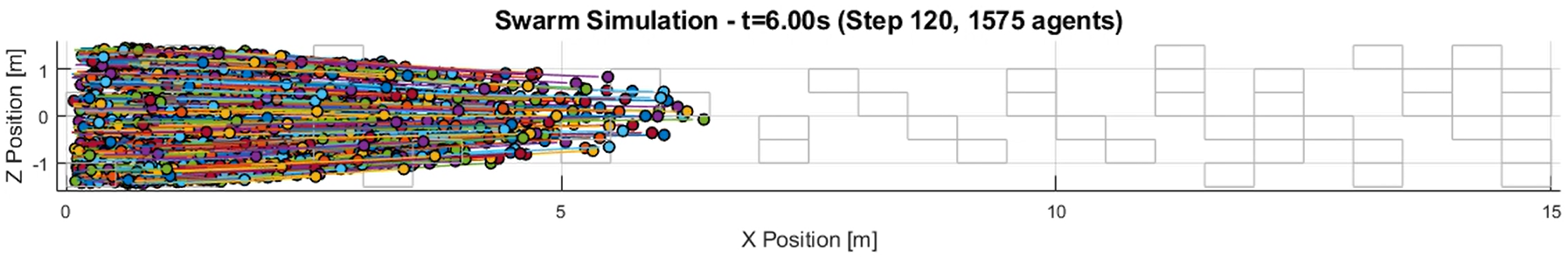}} \\
{\includegraphics[width=0.49\textwidth]{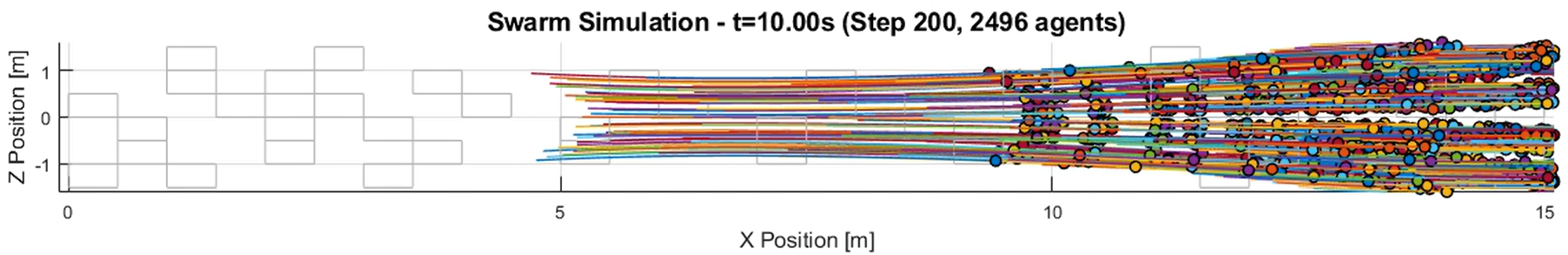}} \hfill
{\includegraphics[width=0.49\textwidth]{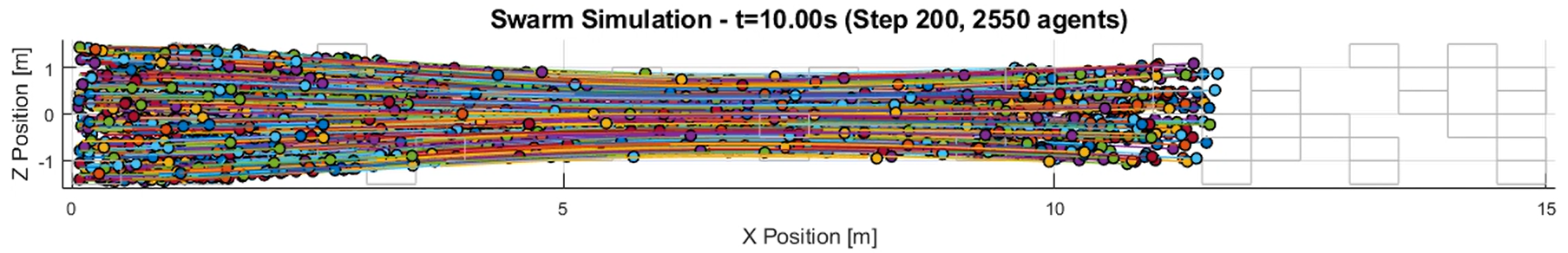}} \\
{\includegraphics[width=0.49\textwidth]{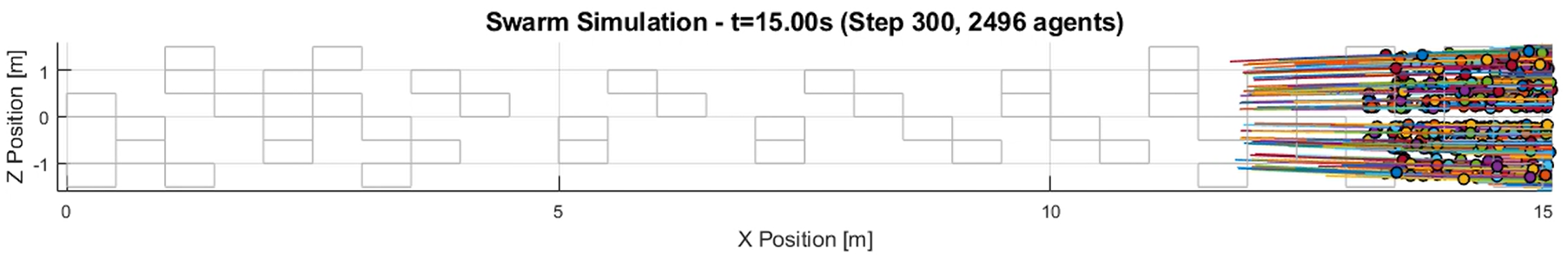}} \hfill
{\includegraphics[width=0.49\textwidth]{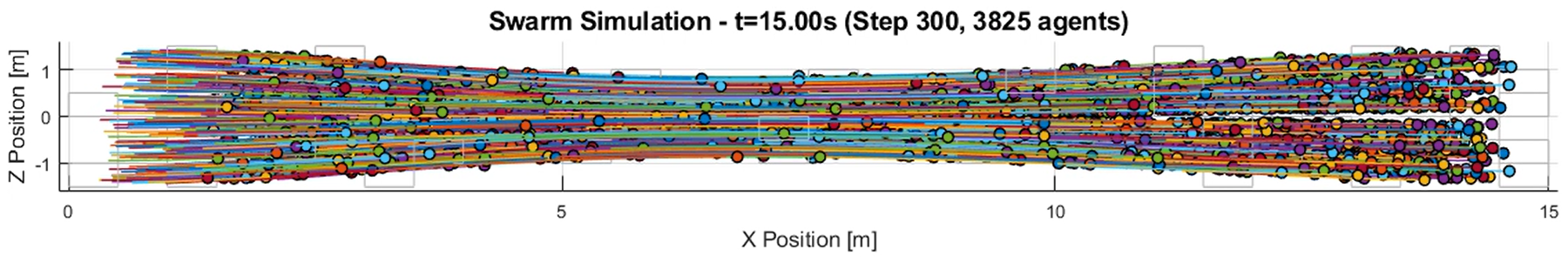}}
\caption{Swarm evolution across two scenarios: Agents seeded in full tunnel (Case 1) and agents admitted at the inlet as a continuous source from a reservoir (Case 2) at time stamps of 0.5 s, 6 s, 10 s, 15 s. Control volume boundaries are shown as gray wireframes.}
\label{fig:swarm_evolution}
\end{figure*}

\section{Swarm Simulation Architecture}
The swarm simulation architecture implements the velocity-fitting described in Section V to derive velocity-only control commands to pass as inputs to the individual agent simulator. Control volume partitioning enables spatial mapping of global control commands to the appropriate agents based on their current location. 

\subsection{Command Mapping}
Each control volume $\Delta V_j(\mathbf{x_{c}})$ of the partitioned tunnel domain, centered at location $\mathbf{x}_{c}$, is assigned a target agent count $N_j^*$ and velocity command $\mathbf{v}_{s,d,j}$ from the velocity-fitting procedure. The agent count $N_j^*$ is initially chosen from a range that is representative of physically realizable agent density, in a manner that allows for the best velocity fit. Once the solution is started, this count is not enforced, and is free to evolve as governed by the physics of the system. At each simulation time step $t_k$, agents are assigned to control volumes based on nearest-center matching:
\begin{equation}
\Delta V_{j,i}(t_k) = \argmin_j \|\mathbf{x}_{i}(t_k) - \mathbf{x}_j\|
\label{eq:cv_assignment}
\end{equation}
Crucially, every agent within $\Delta V_j$ receives the identical velocity command as given in Eq. \ref{eq:scaled_command_map} that is as close as possible to $\mathbf{v}_{s,\text{target},j}$ while respecting quadrotor acceleration/speed limits through the applied scaling factor $S$. 
\begin{equation}
\mathbf{v}_{cmd,i}(\mathbf{x},t) = S* \mathbf{v}_{s,d,j}
\label{eq:scaled_command_map}
\end{equation}
Each agent receives the velocity command $\mathbf{v}_{cmd,i}$ corresponding to its respective $\Delta V_{j}$. This spatially indexed broadcast architecture eliminates the need for inter-agent communication or neighbor sensing. Commands are location-dependent and remain consistent across the swarm, enabling scalability as the swarm size increases.

\subsection{Test Scenarios}
To test a range of swarm-specific scenarios, three different cases were implemented. These vary in initial conditions and agent introduction mechanics but share the same underlying control logic.
\paragraph{Case 1 (Tunnel Seeding)} 
Agents are seeded throughout the tunnel domain according to the tuned agent density profile obtained through velocity-fitting. The agents are populated in all control volumes from inlet to any specific axial section ($x \leq x_{\text{axial}}$) where $N_i^* > 0$, representing static pre-deployment. 
\paragraph{Case 2 (Reservoir)}  
Agents are injected from a reservoir upstream of entry into the tunnel at regular time intervals, simulating persistent generation akin to a fluid inflow process. At each interval $\Delta t_{\text{source}}$, a batch of $N_{\text{batch}}$ agents is introduced at the tunnel inlet with uniform spatial distribution and initial velocity equal to the local field command $\mathbf{v}_{s,\text{entry}}$. The injection rate is tuned to preserve the desired agent density:
\begin{equation}
\frac{N_{\text{batch}}}{\Delta t_{\text{source}}} \approx \frac{N_{\text{entry}}^* \|\mathbf{v}_{s,\text{entry}}\|}{\ell}
\label{eq:source_rate}
\end{equation}

\subsection{Velocity Command Synthesis and Agent Dynamics}
At each time step $\Delta t = 0.05$ s, agents evolve according to their assigned velocity commands and local dynamics. The command synthesis and update loop is shown in Algorithm~\ref{alg:command_loop}.
\begin{algorithm}[h]
\caption{Velocity command execution for each agent}
\label{alg:command_loop}
\begin{algorithmic}[1]
\For{each agent $i$ in a control volume $j$}
    \State Assign control volume
    \Statex \hspace{1.5em}$\Delta V_j \gets \operatorname*{argmin}_{j} \lVert \mathbf{x}_i - \mathbf{x}_j \rVert$
    \State Retrieve command $\mathbf{v}_{\text{cmd},i} \gets \mathbf{v}_{s,d,j}$
    \State Update velocity: \\ \qquad $\mathbf{u}_i \gets \mathrm{VelocityPlant.step}(\mathbf{v}_{\text{cmd},i}, \Delta t)$
    \State Update position: $\mathbf{x}_i \gets \mathbf{x}_i + \mathbf{u}_i \Delta t$
\EndFor
\end{algorithmic}
\end{algorithm}

The combination of global field-driven control and locally enforced physical dynamics ensures that agent trajectories remain dynamically feasible while preserving macroscopic field structure. Real-time velocity and occupancy tracking maintains awareness of agent velocities and densities within each sub-volume, enabling post hoc evaluation of control fidelity and swarm responsiveness.

\subsection{Collision Detection \& Response}
Although the proposed framework does not explicitly model collisions, a lightweight collision-handling scheme was tested to ensure physical plausibility in dense agent scenarios. This addition avoids unrealistic overlaps and enables consistent downstream analysis.
\par
Collisions were detected using proximity checks within spatially partitioned control volumes. When two agents are closer than twice their defined radius, a collision is registered. Interaction types are classified as one-way or mutual based on relative velocities and positions. One-way collisions occur when one agent clearly overtakes another, while mutual collisions involve agents approaching with comparable speed and distance. Mutual cases are further divided into head-on or perpendicular based on their velocity alignment.
\par
A simplified momentum-exchange model adjusts agent speeds without altering direction. In one-way collisions, the slower agent gains speed while the faster one slows down. Mutual interactions result in symmetric energy loss, with greater dissipation in perpendicular approaches. All velocity updates are constrained within defined bounds to prevent numerical instability. This modular collision layer is not central to the proposed contributions but supports a more realistic agent-based environment. It can be extended or replaced as needed for downstream tasks.
\begin{figure}[t]
\centering
\includegraphics[width=\columnwidth]{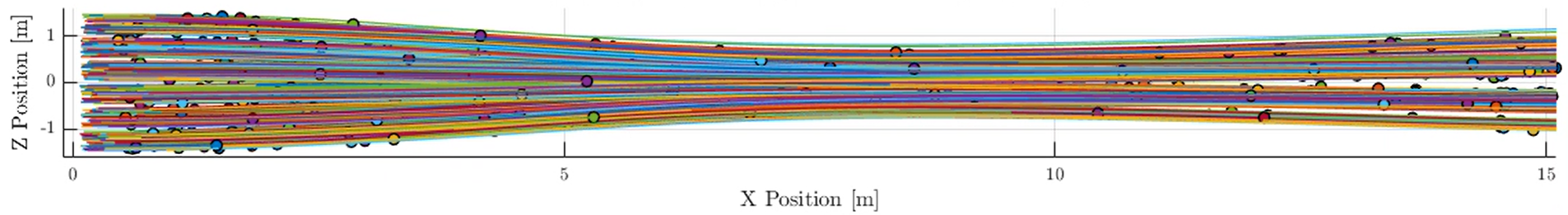} 
\includegraphics[width=\columnwidth]{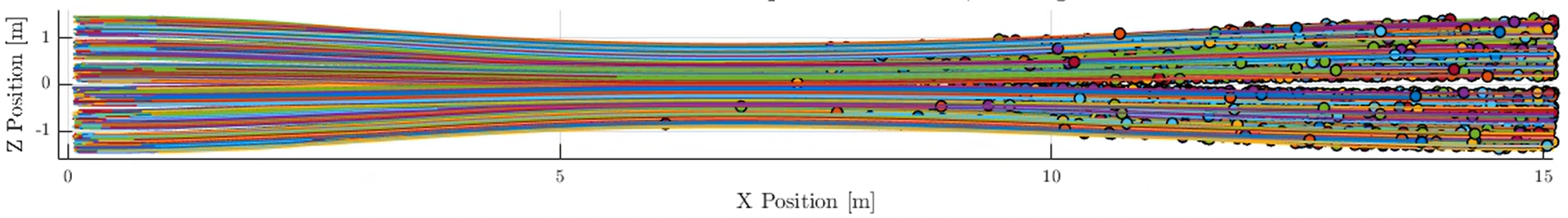} 
\caption{Swarm Visualization plot showing in reservoir mode (Case 1) for (a) Direct CFD unscaled velocities at 5s (top) and (b) velocity-scaled at 20s (bottom).}
\label{fig:vel_comparison_no_scaling}
\end{figure}

\section{Results and Discussion}
For the purpose of this study, the commanded drift velocity $\mathbf{v}_{cmd,i}(\mathbf{x},t)$ is the only specified outer-loop control input, derived based on sub-volume $\Delta V_j$ in which agent $i$ is present at each instant. This choice is made to facilitate future experimental validation with actual low-cost quadcopters which have limited sensory capabilities. Typical outer-loop control input for such robots is the desired velocity state alone. This prescribed velocity field, however, is obtained through the velocity-fitting procedure described in Section~V, where the pressure field \( P_s \) acts as an implicit constraint to ensure correct spatial structure and velocity variance. Consequently, the resulting control is purely velocity-based but encapsulates pressure effects implicitly through fitting. 
\begin{figure}[t]
\centering
\includegraphics[width=\columnwidth]{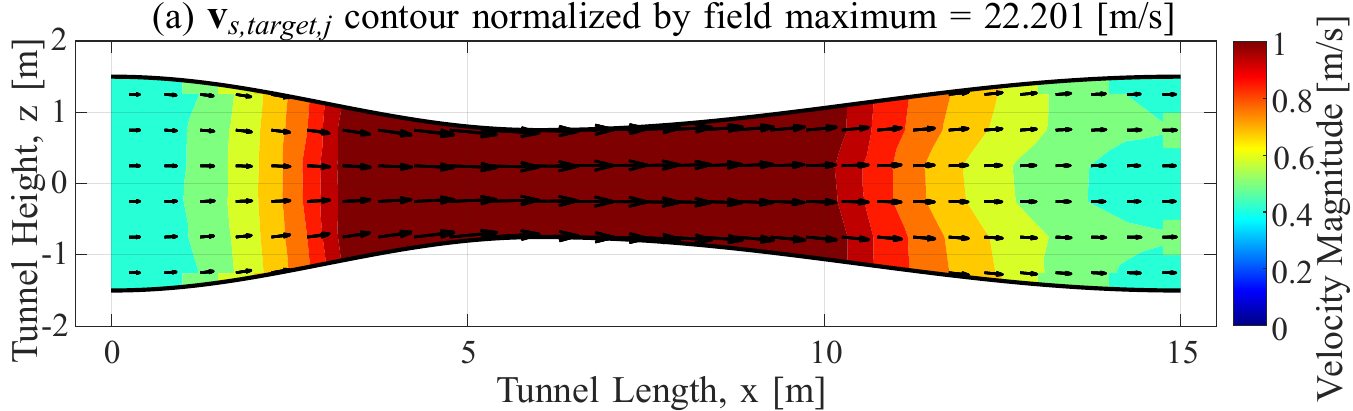}
\includegraphics[width=\columnwidth]{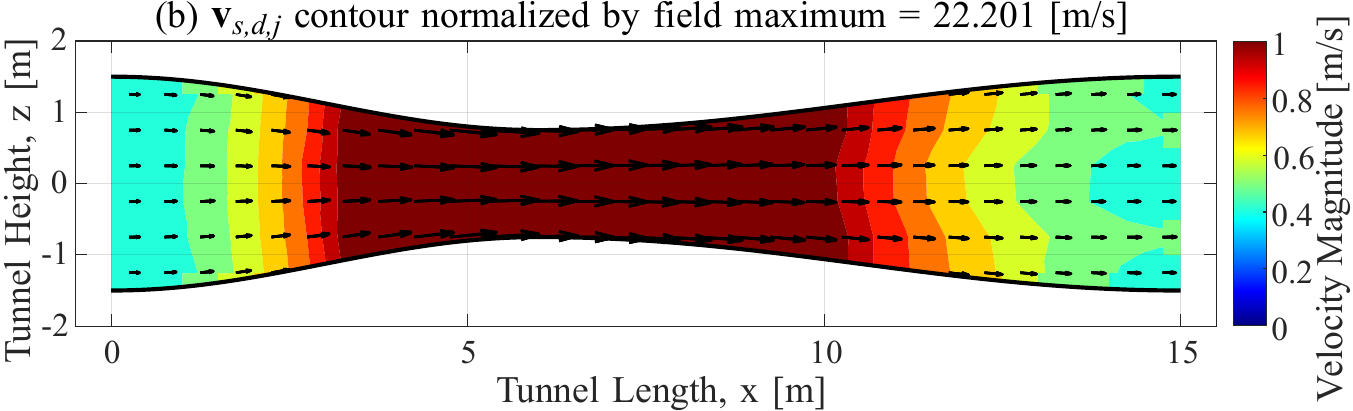}
\includegraphics[width=\columnwidth]{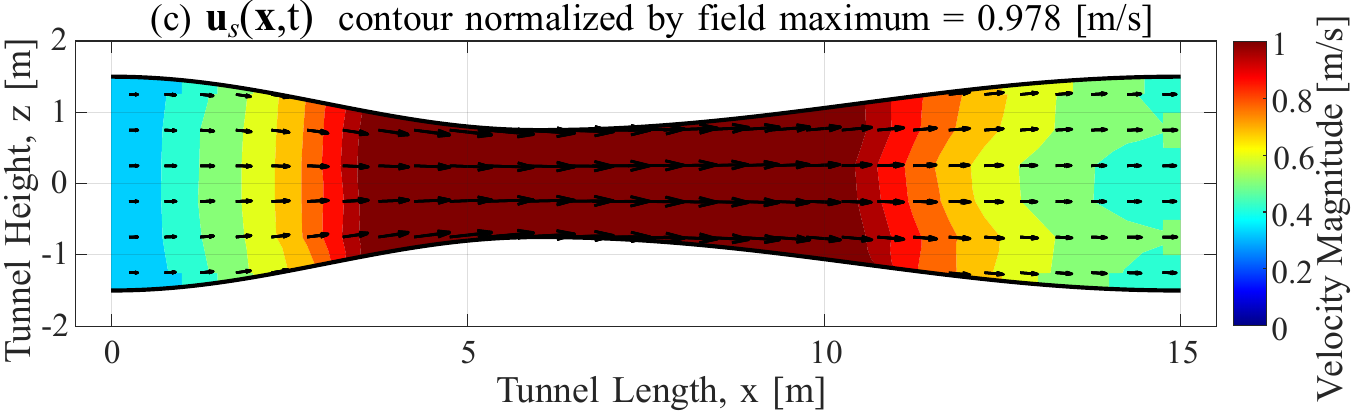}
\caption{Velocity contours showing (a) CFD target velocity field $\textbf{v}_{s,target,j}$, (b) post-velocity-fitting velocity field $\textbf{v}_{s,d,j}$, and (c) swarm simulation-derived velocity $\textbf{u}_{s}(\textbf{x},t)$ contour in reservoir mode (Case 2) in XZ-plane.}
\label{fig:vel_comparison}
\end{figure}

\subsection{CFD Solution}
A Computational Fluid Dynamics (CFD) simulation of airflow through a converging-diverging nozzle provides the reference solution for field-level swarm coordination. Under ideal circumstances, the agents forming the swarm would take on exact values of velocity corresponding to the CFD solution. However, this is not achievable with low-cost physical agents, due to limitations in plant dynamics. 
\par
The CFD simulation yielded steady-state velocity and pressure fields, which were spatially averaged within uniform cubic control volumes of edge length $0.5~m$ resulting in 772 control volumes. The CFD-simulated domain spans $x\in[0,15]$ m, with throat at $x=6~m$. Inlet velocity was set to $3.38~m/s$, producing a peak velocity of $22.4~m/s$ at the throat. This velocity is beyond achievable based on the capabilities of most readily available robots, while still maintaining an acceptable degree of control authority and maneuverability. This required scaling the solution, which is discussed in detail below. 

\subsection{Velocity Fitting and Scaling}
The significance of velocity fitting in the context of velocity-based control was previously discussed. In addition to this, the desired velocity $\textbf{v}_{s,d,j}$ is also scaled to a lower value. This scaling is essential as the compressible fluid velocities (22.4 m/s at the throat) exceed typical physical agent velocity and control authority limits (2-10 m/s). This concern is depicted in Fig. \ref{fig:vel_comparison_no_scaling}, which shows trajectories for individual agents within the swarm when operated in reservoir mode (Case 2). At full velocity, the individual agents are moving too fast in the axial direction and lack actuator authority to replicate the expansion exhibited by the actual fluid. When slowed down to 10\% of the actual velocity, the swarm structure in the latter half of the volume expands in a manner more consistent with the CFD solution to occupy the entire space. 
\par
It is possible to develop a fluid solution with lower velocities, however such a solution corresponds to incompressible flow regime for most fluids and results in a loss in compressibility effects. These effects are highly desirable as they can be leveraged to compensate for limited sensing capabilities of the swarm robots. Implementing the incompressible solution also forces the agents to try and maintain precise distance within each sub-volume (and thus constant overall density), limiting flexibility in structure when navigating cluttered environments. 
\par
\begin{figure}
    \centering
    \includegraphics[width=\columnwidth]{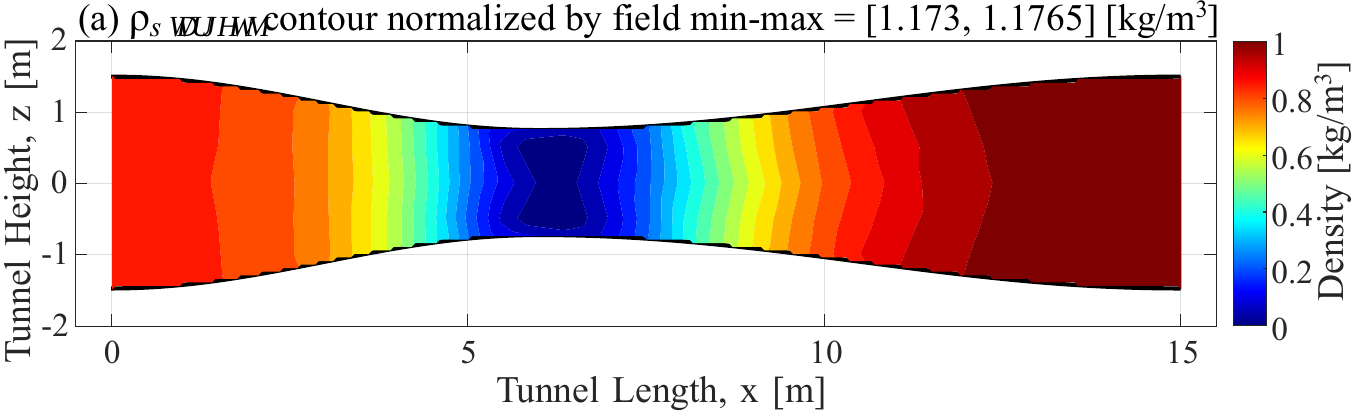}
    \includegraphics[width=\columnwidth]{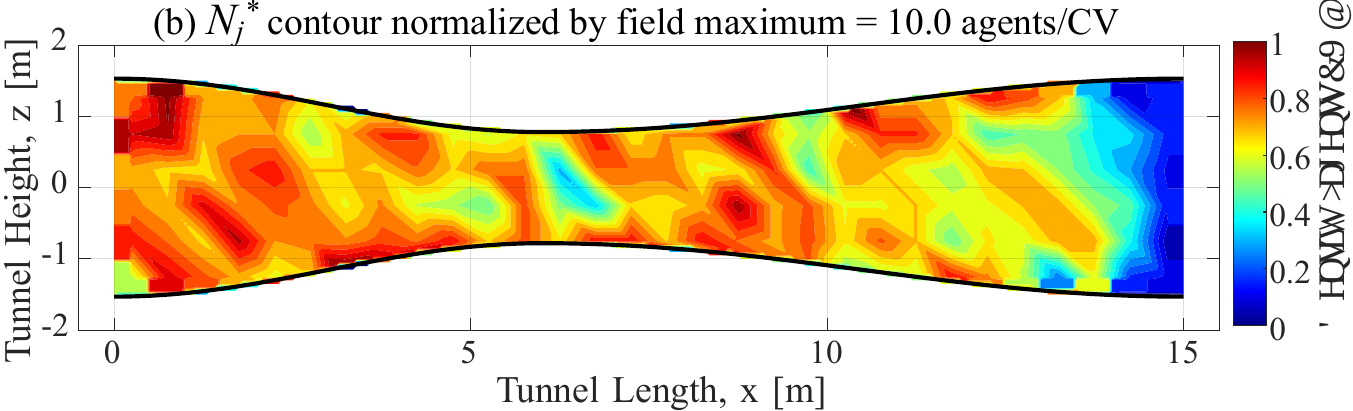}
    \includegraphics[width=\columnwidth]{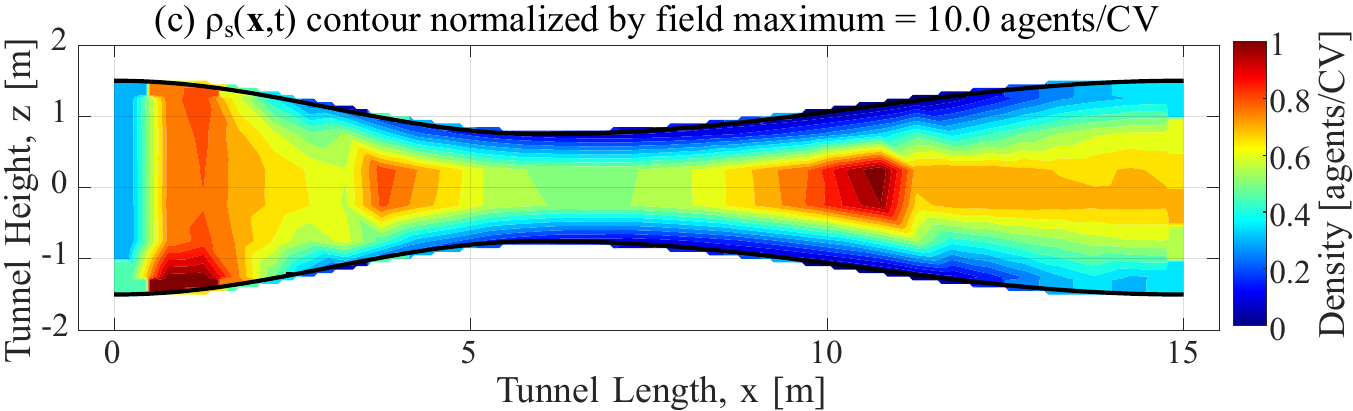}
    \caption{(a) CFD target density field $\rho_{s, target,j}$ and (b) density-scaled ($N^*_j$) plot after velocity fitting and (c) Time-averaged plot of agent density $\rho_{s}(\textbf{x},t)$ in reservoir mode (Case 2) in XZ-plane.}
    \label{fig:density_comparison}
\end{figure}

A significant concern that arises from this scaling is whether it preserves the pressure and velocity values that were computed during the fitting process. The structure of Eqs. \eqref{eq:velocity_constraint} and \eqref{eq:pressure_constraint}, where these quantities are defined for the swarm case, reveals that it is possible to recover identical values of swarm pressure and velocity simply by adjusting the mass and agent number scale factors that appear in these equations. These two parameters result in a different swarm density value than what was previously computed for each location; however, this computed value is only used to seed the velocity fit solution and is never enforced for the actual swarm.
For the case where the swarm behaves similar to how a compressible fluid does, it is expected that all three swarm variables (density, pressure, and velocity) would exhibit identical trends to that of a fluid as it moves through the volume. The density would therefore evolve with trends similar to that of an actual fluid, regardless of the precise numerical value. The remainder of this section thus focuses on presenting and comparing trends in these variables.  This would be the first step in establishing whether the swarm is behaving as a fluid using the proposed definition of fluid equivalent swarm variables. 

\begin{figure}[t]
\centering
\includegraphics[width=\columnwidth]{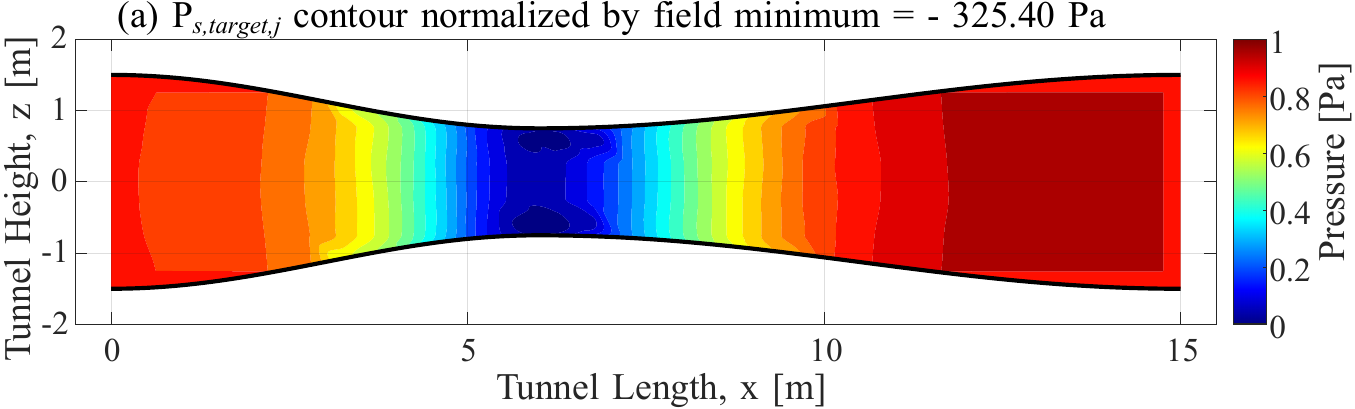}
\includegraphics[width=\columnwidth]{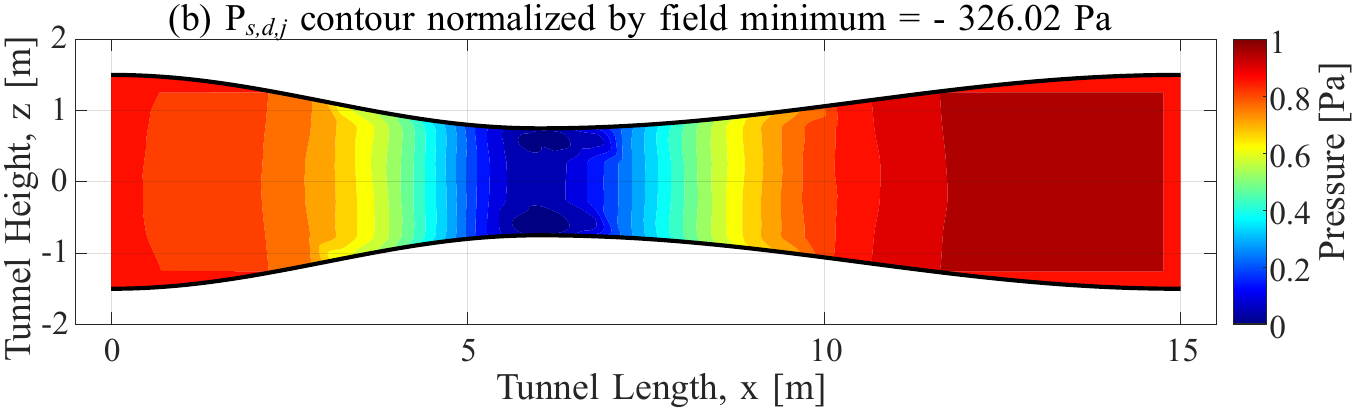}
\includegraphics[width=\columnwidth]{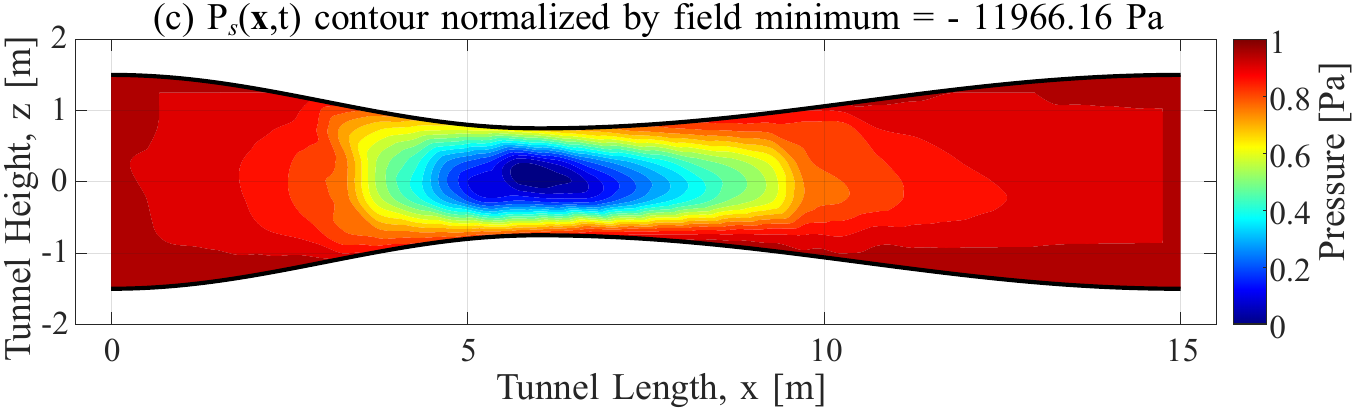}
\caption{Pressure contours showing (a) CFD target field $\textbf{P}_{s,target,j}$, (b) post-velocity-fitting pressure field $P_{s,d,j}$, and (c) simulation-derived pressure $P_{s}(\textbf{x},t)$ contour in reservoir mode (Case 2) in XZ-plane.}
\label{fig:pressure_comparison_sim}
\end{figure}

\subsection{Velocity and Pressure Field Comparison} Initially, three variations of pressure, velocity, and density field values are compared to establish if desired characteristics are exhibited by the collection of agents. The first set comes directly from CFD, the second one is the computed set of variables from the fitting process, and the third is the actual results from implementing the velocity commands to the swarm. Fig. \ref{fig:vel_comparison} shows the normalized velocity results in the XZ-plane cross-section. The results are normalized with respect to the field maximum that occurs around the throat. This is done to focus on the field gradients, and directions which are more informative of whether fluid-like behavior is achieved compared to velocity magnitude. As seen, the fitted velocity closely matches both the magnitude gradients and vector directions of the CFD solution. The flow structure, including the acceleration through the throat and downstream expansion, is reproduced with good fidelity. 
\par
The velocity-fitting respects pressure constraints through velocity variance, which implicitly shapes the emergent swarm density. Thus, the structural coherence arises not from collision handling or containment, but from the $\mathbf{v}_{cmd,i}(\mathbf{x},t)$, which allows the agents to behave collectively like a compressible fluid. Lastly, as seen in Fig. 5 (c) when the velocity command is applied to the swarm after-scaling, the swarm velocity field matches the fluid velocity behavior with good fidelity as well.
\par
The RMSE values of $\textbf{u}_s$ with respect to $\textbf{v}_{s,target,j}$ ranges from 0.15 to 0.9, indicating a generally good alignment, with higher errors in regions of rapid flow changes (e.g., near the throat). Furthermore, these results suggest that velocity-fitting provides the capability for the swarm to adapt to a given environment through tuning of velocity and agent concentrations as desired.
\par
Pressure fields are shown in Fig. \ref{fig:pressure_comparison_sim}. Results in this case are also normalized, but by field minimum occurring near throat. The CFD pressure field shown in Fig. \ref{fig:pressure_comparison_sim} (a) is the static pressure, and is therefore negative with lowest values corresponding to locations in the field that have higher velocity values.  The contours in Fig. \ref{fig:pressure_comparison_sim} (b) and \ref{fig:pressure_comparison_sim} (c) compare post-velocity-fitted pressure and simulation-derived pressure contours using actual agent densities and velocities in the reservoir mode. Also, the results in Fig. \ref{fig:pressure_comparison_sim} (c) are computed from the mean squared difference of the target $\mathbf{v}_{s,target,j}$ and actual velocities $\mathbf{u}_j$ from Eq. \eqref{eq:pressure_constraint}. In addition to qualitative comparison, RMSE values between the normalized values $P_s(\textbf{x},t)$ and $P_{s,target,j}$ ranged from 0 to $0.937$.
The results show that even with velocity-only control the important aspects of the pressure field behavior are being retained in the swarm domain. 
Higher discrepancies were observed near the converging and diverging sections of the domain, where sharp gradients in the discrete swarm motion induced increased local acceleration demands. Despite these localized deviations, the overall pressure structure including compression at the throat and expansion downstream was consistently preserved. This indicates that the proposed pressure definition captures the correct macroscopic behavior required to maintain swarm coherence under finite population effects and decentralized execution. It is worth noting that pressure in Fig. \ref{fig:vel_comparison} (c) was not computed using density values that were obtained from the optimization performed during the fitting procedure, but instead using the time-averaged density of agents present at a given location. This further suggests that the choice of swarm domain variables results in a fluid-like motion of the collection of agents.

\begin{figure}
    \centering
    \includegraphics[width=1\linewidth]{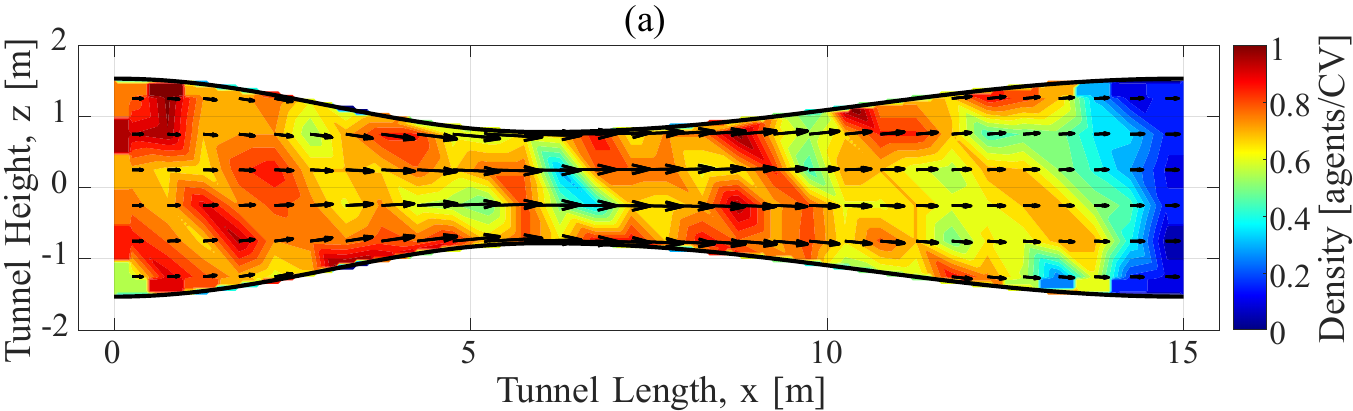}
    \includegraphics[width=1\linewidth]{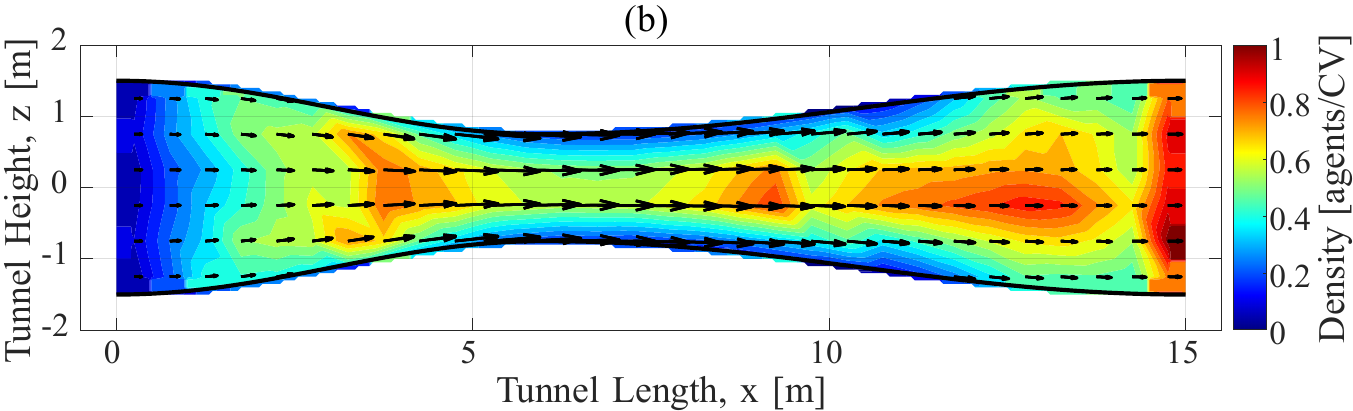}
    \includegraphics[width=1\linewidth]{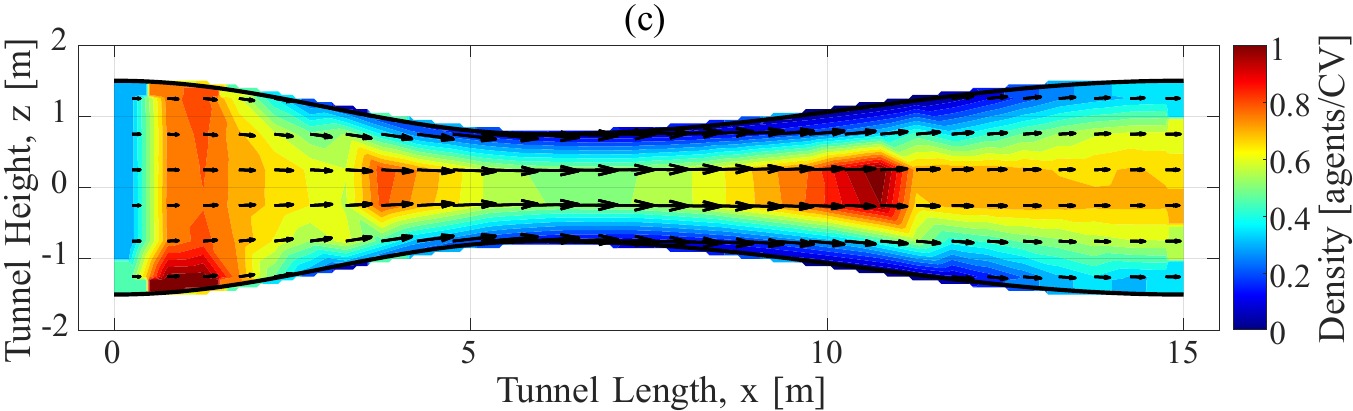}
    \includegraphics[width=1\linewidth]{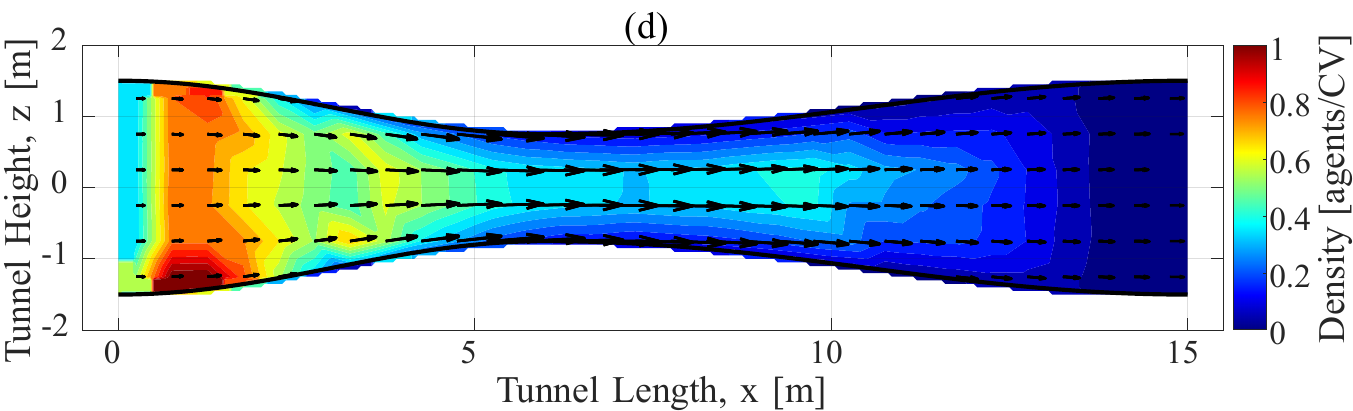}
    \caption{Time-Averaged contour plots of agent density overlaid on velocity vector field in XZ plane with the vector size representative of the velocity magnitude. The plots show scenarios of velocity-fitted (a) Control command field, (b) time-averaged field of Case 1 with agents seeded from 0-3m, (c) time-averaged field of Case 2 for 60s without collisions, and (d) time-averaged field of Case 2 with collisions.}
    \label{fig:density_velocityfield_time_averaged}
\end{figure}

\subsection{Density Evolution and Specific Quantitative Comparisons}
The density variable for this framework is the most significant as it is seeded as opposed to continually specified. If the swarm is truly behaving like a compressible fluid, then it is reasonable to expect that the density would exhibit similar trends as the swarm moves through the volume. Among the two tested scenarios, the reservoir-based continuous inflow case (Case 2) uniquely illustrates the time-evolving behavior of the swarm as it propagates through the domain. Unlike tunnel seeding (Case 1), where the swarm is initialized across the full volume, the reservoir case simulates a sustained inflow of agents from the inlet. This setup enables detailed observation of how the swarm self-organizes over time under velocity-only control, and is therefore the subject of focus for quantitative comparison.
\par
The density field values for CFD, velocity-fitting, and the implemented case are shown in Fig. \ref{fig:density_comparison}. It is immediately obvious that the density values that are used to seed the velocity fitting procedure, shown in Fig. \ref{fig:density_comparison} (b), are substantially different from the desired (CFD) density values shown in  Fig. \ref{fig:density_comparison} (a). However, the actual average density value as the swarm field evolves in time, as shown in Fig. \ref{fig:density_comparison} (c), exhibits the characteristic trend of compressible fluid being driven through this contour at subsonic speeds. The density is higher at the inlet, and decreases slightly as the agents speed up to clear the throat, after which they expand slowing down into a higher density value towards the exit.
\par
The results shown in Fig. \ref{fig:density_velocityfield_time_averaged} substantiate this behavior for both Case 1 and Case 2 of agent seeding. For this case, Fig. \ref{fig:density_velocityfield_time_averaged} (b) and (c), show that regardless of the value used for fitting the velocity (shown in Fig. \ref{fig:density_velocityfield_time_averaged} (a)), the density field evolves in a manner akin to the CFD flow field enhancing confidence in the selection of the primitive equivalent variables used to represent the swarm. Furthermore, from a quantitative perspective, the RMSE values of $\rho_s$ w.r.t. $\rho_{s, target, j}$ ranges from 0.61 to 0.98, reflecting some variation between time-averaged discrete agent concentrations in reservoir scenario (Case 2) and fluid density. The focus however remains on trends due to the fact that the proper equivalent of fluid-properties in the swarm domain requires further analysis that is part of future work.
\par
\begin{figure*}[htpb]
    \centering
    {\includegraphics[width=0.32\textwidth]{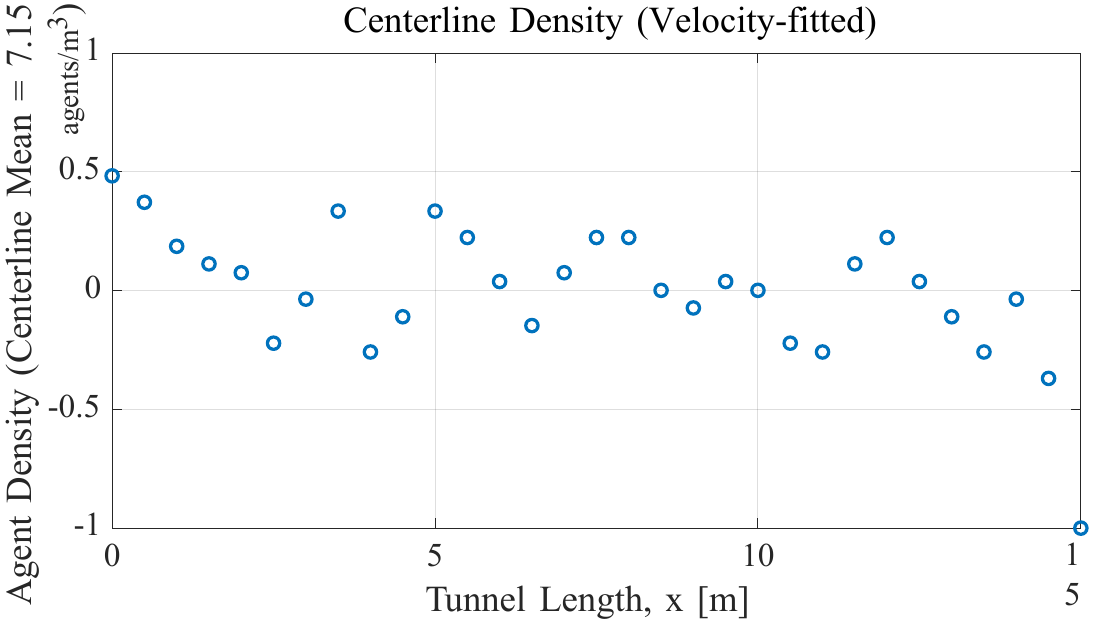}} \hfill
    {\includegraphics[width=0.32\textwidth]{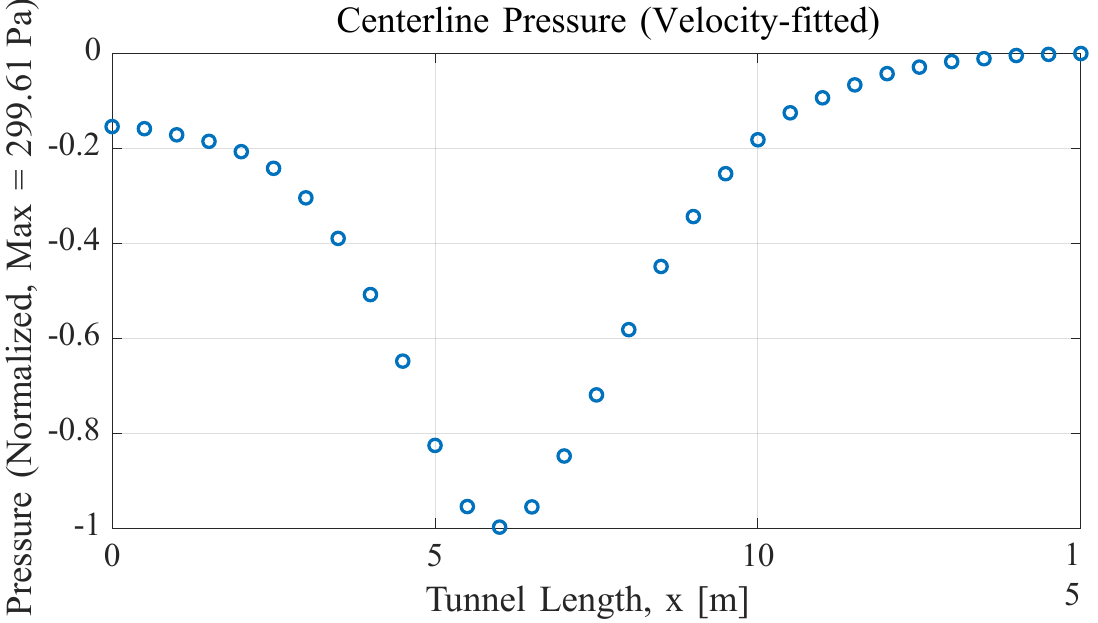}} \hfill
    {\includegraphics[width=0.32\textwidth]{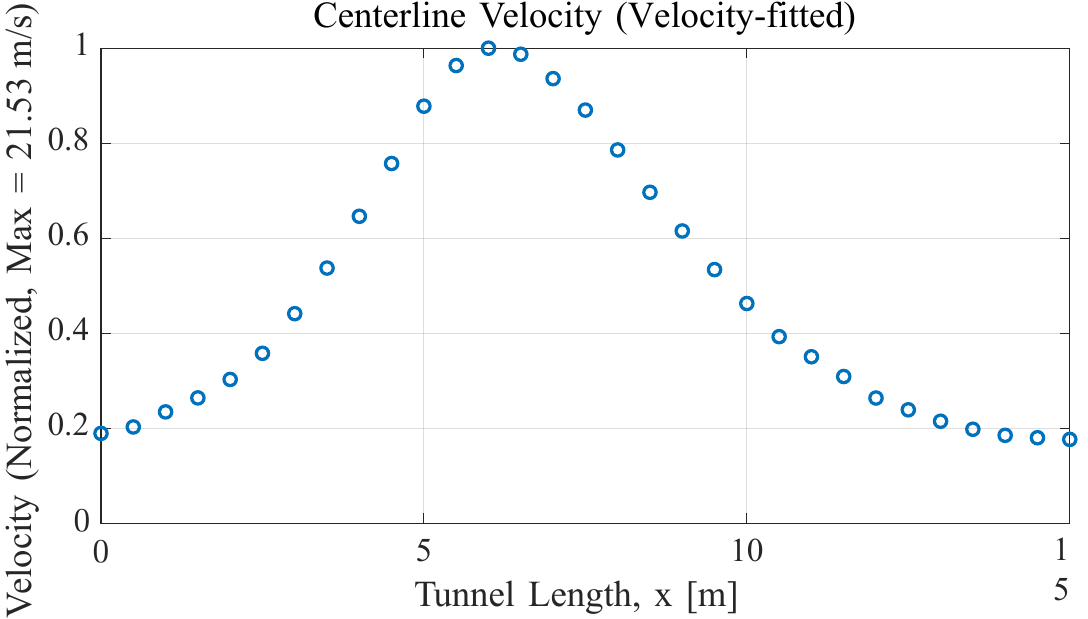}} \\
    {\includegraphics[width=0.32\textwidth]{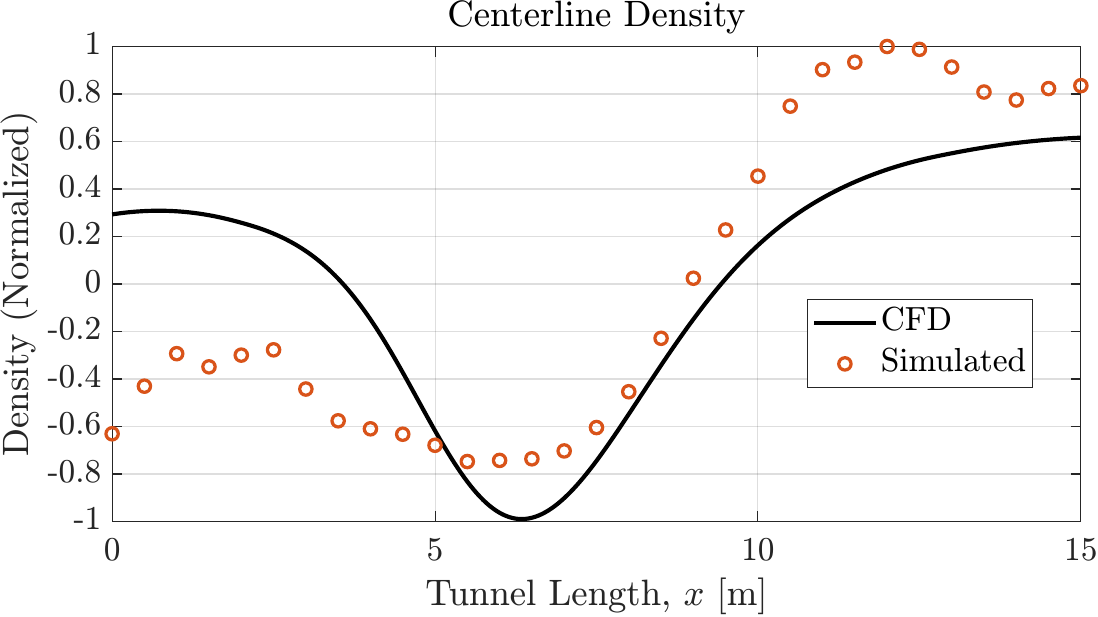}} \hfill
    {\includegraphics[width=0.32\textwidth]{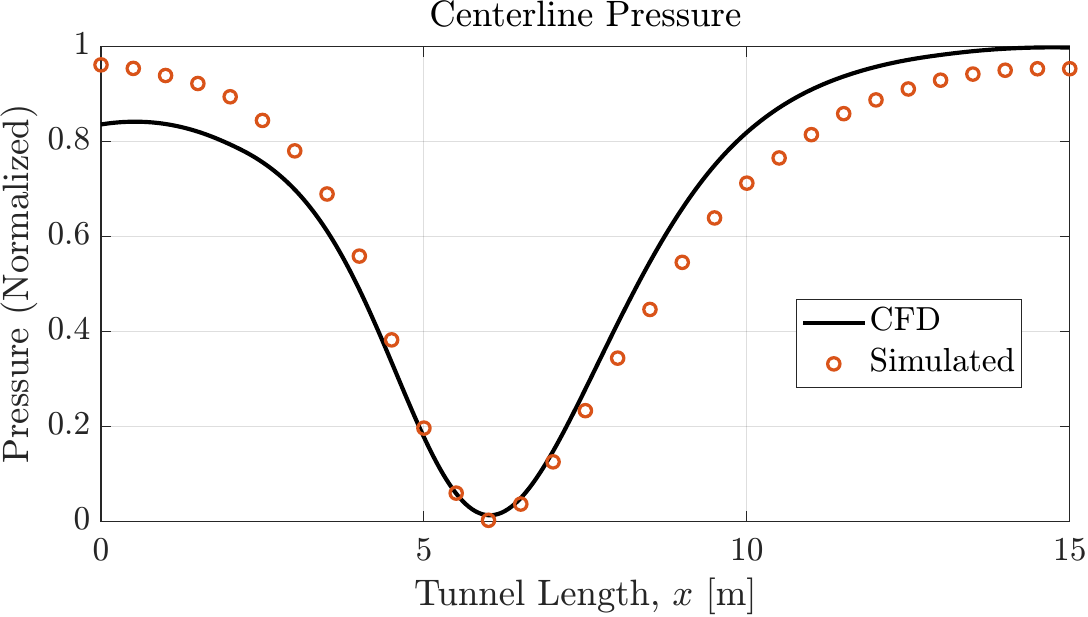}} \hfill
    {\includegraphics[width=0.32\textwidth]{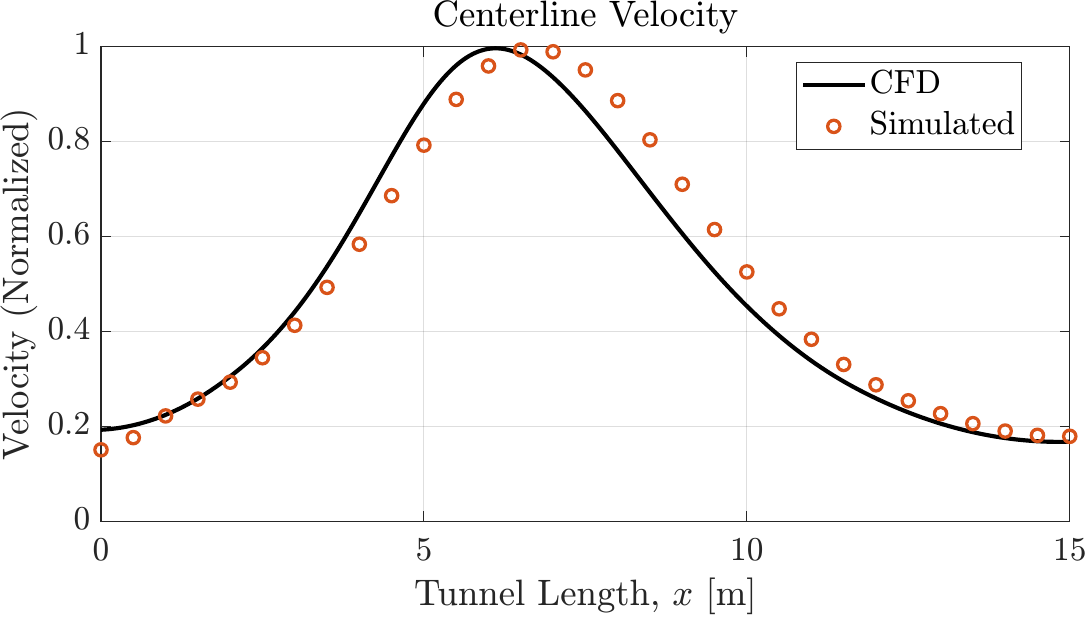}} \\
    \caption{Centerline Plots for the reservoir scenario (Case 2) run for 300 seconds.}
    \label{fig:centerline_plots}
\end{figure*}

Lastly, a more quantitative comparison between the density, pressure and velocity is presented in Fig. \ref{fig:centerline_plots}. For this instance, the normalized pressure, velocity and the normalized density variation from field maximum, was compared between the CFD case and the case with simulated agents. The pressure and velocity trends match for both cases with good fidelity of \(\approx\)10\% or less. While the density trends have more variation, they behave very much like that of a compressible fluid, even when seeded with values that were substantially different.   

\subsection{Collisions}
The impact of potential collisions is studied through a simplified collision scheme discussed in the Methods sections, and is characterized for the reservoir mode (Case 1). Key metrics such as average velocity profiles, density distributions, and overall structure remained similar between the two cases with and without collisions.  Results for density evolution with collisions are shown in Fig. \ref{fig:density_velocityfield_time_averaged} (d). These results indicate a similar trend (density is higher at the inlet, reduces at the throat, and increases as the swarm heads towards the exit) compared to the collision-free case. However, for this case swarm flow takes longer to evolve as the collision mitigation strategy is competing against the commanded velocity. As a result, the precise values for the collision and collision-free case differ, however, the trends are still consistent with that of a real fluid.   

\section{Summary}
This work presents a fluid-inspired framework for decentralized swarm control based on a minimal quartet of primitive state variables analogous to those used in classical fluid dynamics. The four primitive variables \( Q = \{\rho_s, u_s, T_s, P_s\} \), were derived directly from agent-level motion analogous to how fluid elements move. While all four variables were fully defined, only the velocity field,  $\textbf{v}_{cmd,i}(\textbf{x},t) $, was used for control execution. This field is derived through a velocity-fitting procedure that embeds pressure \( P_s \) implicitly by matching velocity variance to target pressure, enabling purely velocity-based control to reproduce fluid-consistent swarm structure without explicit boundary enforcement or inter-agent coordination. The four variables; swarm velocity, density, pressure, and temperature were derived directly from agent-level motion and formulated to preserve the structural role of conservation laws at finite swarm sizes.
\par
A velocity-fitting procedure was introduced to generate swarm-consistent velocity-only commands $\textbf{v}_{cmd,i}(\textbf{x},t) $  that implicitly reflect pressure effects through velocity variance. Simulation results across multiple deployment scenarios demonstrate that this velocity-only execution is sufficient to maintain global coherence. They reproduce compressible flow features, and preserve spatial structure over time, even under continuous agent inflow and without inter-agent communication or physical boundary enforcement. This separation between variable formulation and control execution ensures that the framework can be incrementally expanded as sensing and computational capabilities increase.
\par
The remaining primitive variables \( \rho_s \) and \( T_s \) were computed and monitored but not actively used for control. The former was allowed to evolve in response to the applied commands and compared to the CFD case to demonstrate that this formulation of variables does capture important fluid flow features. Lastly, the formulation of these two variables provides a foundation for future extensions in which density regulation, pressure-aware shaping of motion, and/or stability-related adaptation will be incorporated. \( T_s \), as a measure of control uncertainty and local disorder, provides a pathway for stability-aware adaptation and energy-aware planning. Such extensions are expected to become important in environments involving obstacles, disturbances, or heterogeneous agent capabilities.
\par
Overall, the results confirm that a small set of fluid-equivalent primitive variables can support scalable, decentralized swarm motion that exhibits flow-like structure consistent with continuum behavior. The framework establishes a pathway for extending swarm coordination strategies using principles drawn from fluid mechanics while maintaining compatibility with discrete, finite populations of robotic agents.

\section{Conclusions and Future Work}
This work presents a fluid-inspired framework for decentralized control of robotic swarms, grounded in a quartet of primitive variables analogous to those in fluid mechanics. Simulations across multiple test scenarios demonstrated that a swarm structure and spatial coherence analogous to a collection of fluid elements that form a fluid body can emerge even from pure velocity control. Temporal results confirmed that structure is preserved even under continuous agent inflow, validating the use of velocity-fitted values as the sole control commands.
\par
While only the velocity field is used for control in this study, the full primitive variable set offers a foundation for future extensions. Incorporating explicit pressure, density, and temperature control could enable responsive behavior in obstacle-rich or uncertain environments. The proposed framework thus offers a scalable, physically consistent approach to collective motion, capable of bridging analytical models and real-world swarm robotics.
\par
This study is intended to be a platform-agnostic foundation for a new way to coordinate large swarms of robotic agents. While certain hardware characteristics of the swarm determine the fluid-equivalent swarm properties, they need to be established only once for a collection of robots and may generalize to certain classes of robots. Many potential future directions exist; however, current efforts are focused on understanding the density parameter for agents that have similar dynamic characteristics. For the analysis presented herein, a rudimentary collision mitigation system was investigated. As the immediate next step, a collision system that has uncertainty on par with state-of-the-art hardware shall be implemented. A relationship between the uncertainty and the swarm density parameter shall be established analytically. The aim is to develop a strategy to control the density (through velocity) such that it can account for location uncertainty to minimize the likelihood of collisions and allow for the robots to "flow" more akin to an actual fluid.
\par
Lastly, the implementation of this architecture on an actual swarm would require the ability to compute a fluid solution similar to CFD for each agent. A complete solution similar to the one used in this study would be too expensive to implement on low-cost commercial hardware, however, learning based solutions can be implemented well within these constraints. Thus, each agent can efficiently compute how it needs to operate. A common model is used across all agents that requires only generic fluid and geometric boundary conditions. The former comes from user-specified input based on the desired direction and structure, whereas the latter can be specified or developed using rudimentary sensors that are part of most modern robots. This enables robust decentralization without excessive expense.       

\section{Nomenclature}
\label{sec:nomenclature}
\vspace{-0.5em}

\begin{table}[H]
\footnotesize
\centering
\renewcommand{\arraystretch}{0.92}
\setlength{\tabcolsep}{3pt}
\setlength{\textfloatsep}{6pt}
\setlength{\intextsep}{6pt}
\begin{tabularx}{\columnwidth}{@{}l >{\RaggedRight\arraybackslash}X@{}}
\toprule
\textbf{Symbol} & \textbf{Description} \\
\midrule
Subscript $s$ & Swarm-field (macroscopic) variable \\
Subscript $d$ & Desired or prescribed variable \\
$i$ & Agent index \\
$j$ & Control-volume index \\
$t$ & Time [s] \\
$P$ & Pressure of an isentropic compressible gas [Pa] \\
$T$ & Thermodynamic temperature (or energy-per-mass surrogate used in this work) [m$^2$/s$^2$] \\
$\rho$ & Density of an isentropic compressible gas [kg/m$^3$] \\
$\mathbf{u}$ & Velocity vector of an isentropic compressible gas [m/s] \\
$\gamma$ & Specific heat ratio $c_p/c_v$ \\
$k$ & Barotropic constant in $P = k\,\rho^{\gamma}$ \\
$P_s$ & Swarm pressure analogue [Pa] \\
$T_s$ & Swarm temperature analogue [m$^2$/s$^2$] \\
$\rho_s$ & Swarm density (agent concentration) [agents/m$^3$]  \\
$\mathbf{u}_s$ & Swarm bulk velocity vector [m/s] \\
$Q$ & Quartet: swarm primitive state-variable set $Q=\{\mathbf{u}_s,\rho_s,P_s,T_s\}$ \\
$\mathbf{u}_{d}(\mathbf{x},t)$ & Desired bulk drift velocity field [m/s] \\
$\hat{u}_d$ & Unit vector in the desired drift direction \\
$\mathbf{x}$ & Spatial position vector $\mathbf{x}=(x,y,z)$ [m] \\
$\mathbf{v}_{\mathrm{CFD}}$ & CFD velocity vector at grid nodes [m/s] \\
$P_{\mathrm{CFD}}$ & CFD static pressure [Pa] \\
$\mathbf{v}_{s,\mathrm{target},j}$ & Target mean velocity in control volume $j$ [m/s] \\
$P_{s,\mathrm{target},j}$ & Target mean pressure in control volume $j$ [Pa] \\
\bottomrule
\end{tabularx}
\end{table}
\begin{table}[H]
\footnotesize
\centering
\renewcommand{\arraystretch}{0.92}
\setlength{\tabcolsep}{3pt}
\setlength{\textfloatsep}{6pt}
\setlength{\intextsep}{6pt}
\begin{tabularx}{\columnwidth}{@{}l >{\RaggedRight\arraybackslash}X@{}}
\toprule
\textbf{Symbol} & \textbf{Description} \\
\midrule
$\rho_{s,\mathrm{target},j}$ & Target mean density in control volume $j$ [kg/m$^3$] \\
$m_i$ & Mass of agent $i$ [kg] \\
$\mathbf{x}_i(t)$ & Position of agent $i$ [m] \\
$\mathbf{u}_i(t)$ & Realized velocity of agent $i$ [m/s] \\
$\mathbf{v}_{s,d,j}$ & Desired swarm velocity in control volume $j$ [m/s] \\
$P_{s,d,j}$ & Desired swarm pressure in control volume $j$ [Pa] \\
$\mathbf{v}_{\mathrm{cmd},i}(\mathbf{x},t)$ & Commanded velocity sent to agent $i$ based on its current $\Delta V_j$ at time $t$ [m/s] \\
$S$ & Scaling factor applied to commanded velocities \\
$V$ & Total tunnel domain volume [m$^3$] \\
$x_{axial}$ & Axial distance from domain entry [m] \\
$\Delta V$ & Cubic control-volume size [m$^3$] \\
$\ell$ & Control-volume edge length [m] \\
$\mathbf{x}_c$ & Control-volume center location $\mathbf{x}_c=(x_c,y_c,z_c)$ [m] \\
$\Delta x$ & Axial slice length along $\hat{u}_d$ [m] \\
$A$ & Cross-sectional area normal to $\hat{u}_d$ [m$^2$] \\
$N_j$ & Number of agents in control volume $j$ \\
$N_j^{*}$ & Optimized number of agents in control volume $j$ \\
$N_{\min},\,N_{\max}$ & Minimum/maximum agents per control volume \\
$K_{\mathrm{iter}}$ & Average iterations per candidate $N_j$ \\
$\bar{\mathbf{u}}$ & Mean candidate velocity during optimization [m/s] \\
$\alpha$ & Velocity–pressure error weighting \\
$\mathcal{L}$ & Optimization loss function \\
$\boldsymbol{\xi}_j$ & Gaussian initialization perturbation \\
$\sigma$ & Standard deviation of perturbations \\
$\epsilon$ & Convergence tolerance \\
$L_c$ & Characteristic inter-agent spacing [m] \\
$c_{v,s}$ & Swarm specific heat at constant volume \\
$c_{p,s}$ & Swarm specific heat at constant pressure \\
$R_s$ & Swarm gas constant $R_s=c_{p,s}-c_{v,s}$ \\
$\gamma_s$ & Swarm specific heat ratio $\gamma_s=c_{p,s}/c_{v,s}$ \\
$k_s$ & Swarm barotropic constant in $P_s=k_s\,\rho_s^{\gamma_s}$ \\
$c$ & Swarm speed-of-sound analogue [m/s] \\
$\omega$ & Angular frequency of disturbances [rad/s] \\
$\tau$ & Actuation / reaction delay [s] \\
$\mathbf{U}$ & Local mass-weighted mean agent velocity [m/s] \\
$K_{b,s}$ & Swarm-equivalent Boltzmann constant \\
$\mathbf{a}_T$ & Thrust acceleration vector [m/s$^2$] \\
$\mathbf{F}_d$ & Aerodynamic drag force [N] \\
$\tau_T$ & Thrust dynamics time constant [s] \\
$a_{\max}$ & Maximum achievable specific acceleration [m/s$^2$] \\
$T_{\max}$ & Maximum thrust force [N] \\
$\theta_{\max}$ & Maximum body tilt angle [rad] \\
$g$ & Gravitational acceleration [m/s$^2$] \\
$\Delta t$ & Simulation time-step size [s] \\
$t_k$ & Discrete time-step index \\
$N_{\mathrm{source}}$ & Total injected agents in reservoir mode \\
$N_{\mathrm{batch}}$ & Agents injected per batch \\
$N^*_{\mathrm{entry}}$ & Time-varying injected agent count \\
$\Delta t_{\mathrm{source}}$ & Time between batch injections [s] \\
$\mathbf{v}_{s,\mathrm{entry}}$ & Entry command velocity [m/s] \\

\bottomrule
\end{tabularx}
\end{table}

\section*{Acknowledgments}
The authors acknowledge the J. Mike Walker ’66 Department of Mechanical Engineering at Texas A\&M University for providing the academic environment, computational resources, and research support that made this work possible.

\section*{Disclosure Statement}
The authors report no conflicts of interest. AI-assisted tool (OpenAI's ChatGPT 5.2) was used to improve the clarity and readability of the manuscript. All content was reviewed and approved by the authors, who assume full responsibility for the final published version.

\section*{Data Availability Statement}
The datasets and code used for this study are available from the corresponding author upon reasonable request.

\section*{ORCID}
\noindent Mohini Priya Kolluri \orcidlink{https://orcid.org/0009-0006-3829-6005} \href{https://orcid.org/0009-0006-3829-6005}{https://orcid.org/0009-0006-3829-6005} \\
Ammar Waheed \orcidlink{https://orcid.org/0000-0001-5042-6593} \href{https://orcid.org/0000-0001-5042-6593}{https://orcid.org/0000-0001-5042-6593}\\
Zohaib Hasnain \orcidlink{https://orcid.org/0000-0002-8358-451X} \href{https://orcid.org/0000-0002-8358-451X}{https://orcid.org/0000-0002-8358-451X}

\bibliographystyle{IEEEtran}
\bibliography{References}

\end{document}